\RequirePackage{fix-cm}
\documentclass[twocolumn]{svjour3} 
\smartqed  
\usepackage{graphicx}
\usepackage{url}
\usepackage{array}
\usepackage{amsmath}
\usepackage{float}
\usepackage{color}

\usepackage{booktabs}
\usepackage{multirow} 
\usepackage{ulem}
\usepackage{soul}
\usepackage{algorithm}
\usepackage{algorithmic}
\usepackage{amssymb}
\usepackage{booktabs}
\usepackage{bbding}
\usepackage{overpic}
\usepackage{subfigure}
\usepackage{makecell}
\usepackage{array}
\usepackage{colortbl,xcolor}
\usepackage{amssymb}
\usepackage{subfigure}
\usepackage{colortbl,xcolor}
\usepackage{pifont}
\usepackage[breaklinks=true,colorlinks,citecolor=blue,urlcolor=blue,linkcolor=blue,bookmarks=false,pagebackref=true]{hyperref}

\newcolumntype{M}[1]{>{\centering\arraybackslash}m{#1}}

\definecolor{bblue}{rgb}{0,150,230}
\definecolor{mygray}{gray}{.9}
\definecolor{lightgray}{gray}{.96}
\definecolor{myy}{RGB}{126,95,0}
\definecolor{ggray}{RGB}{127,127,127}
\definecolor{mygreen}{RGB}{93,173,85}
\definecolor{myred}{RGB}{240,16,89}
\definecolor{myblue}{RGB}{0,114,188}
\definecolor{darkgreen}{rgb}{0.0, 0.5, 0.0}
\definecolor{demphcolor}{RGB}{100,100,100}
\newcolumntype{C}[1]{>{\centering\let\newline\\\arraybackslash\hspace{0pt}}m{#1}}

\DeclareRobustCommand\onedot{\futurelet\@let@token\@onedot}
\newcommand{\etal}{\textit{et al}. }
\newcommand{\ie}{\textit{i.e., }}
\newcommand{\eg}{\textit{e.g., }}

\makeatother

\newcolumntype{d}[1]{>{\raggedright\arraybackslash}p{#1pt}}
\newcolumntype{e}[1]{>{\raggedleft\arraybackslash}p{#1pt}}

\newcommand{\cmark}{\ding{51}}%
\newcommand{\xmark}{\ding{55}}%
\mathchardef\mhyphen="2D
\graphicspath{{image/}}
\begin{document}

\title{GridFormer: Residual Dense Transformer with Grid Structure for Image Restoration in Adverse Weather Conditions
}


\author{Tao Wang         \and
        Kaihao Zhang        \and
        Ziqian Shao        \and
        Wenhan Luo  \and \\
        Bjorn Stenger  \and 
        Tong Lu   \and
        Tae-Kyun Kim  \and
        Wei Liu \and
        Hongdong Li 
}

\institute{
         Tao Wang, Ziqian Shao and Tong Lu  \at
         Nanjing University, Nanjing, 210023, China \\
         \email{taowangzj@gmail.com, ziqian.shao@outlook.com, lutong@nju.edu.cn}        \\
         \and
        Kaihao Zhang \at Harbin Institute of Technology, Shenzhen, 518055, China \\
        \email{super.khzhang@gmail.com}\\
        \and
        Wenhan Luo \at
        Hong Kong University of Science and Technology, Hong Kong\\
        \email{whluo.china@gmail.com} \\
        \and
         Bjorn Stenger \at Rakuten Institute of Technology, Japan \\
           \email{bjorn@cantab.net} \\
        \and
        Tae-Kyun Kim \at Imperial College London, London, UK \& KAIST, Daejeon, South Korea\\
           \email{tk.kim@imperial.ac.uk} \\  \and
        Wei Liu \at Tencent, Shenzhen, 518107, China \\
           \email{wl2223@columbia.edu} \\
                   \and
        Hongdong Li \at
        Australian National University, Australia\\
        \email{hongdong.li@gmail.com}  \\
}

\date{Received: date / Accepted: date}

\maketitle

\begin{abstract}
Image restoration in adverse weather conditions is a difficult task in computer vision. In this paper, we propose a novel transformer-based framework called GridFormer which serves as a backbone for image restoration under adverse weather conditions. GridFormer is designed in a grid structure using a residual dense transformer block, and it introduces two core designs. First, it uses an enhanced attention mechanism in the transformer layer. The mechanism includes stages of the sampler and compact self-attention to improve efficiency, and a local enhancement stage to strengthen local information. Second, we introduce a residual dense transformer block (RDTB) as the final GridFormer layer. This design further improves the network's ability to learn effective features from both preceding and current local features. The GridFormer framework achieves state-of-the-art results on five diverse image restoration tasks in adverse weather conditions, including image deraining, dehazing, deraining \& dehazing, desnowing, and multi-weather restoration. The source code and pre-trained models are available at \url{https://github.com/TaoWangzj/GridFormer}.
\end{abstract}

\section{Introduction}\label{section:Introduction}

Capturing high-quality images in adverse weather conditions like rain, haze, and snow is a challenging task due to the complex degradation that occurs in such conditions. These include color distortion, blur, noise, low contrast, and other issues that directly lower the visual quality. Furthermore, such degradation can lead to difficulties in downstream computer vision tasks such as object recognition and scene understanding~\cite{itti1998model,carion2020end}.

Traditional methods for image restoration in adverse weather conditions often rely on handcrafted priors such as smoothness and dark channel, with linear transformations~\cite{roth2005fields,garg2005does,he2010single,chen2013generalized}. However, these methods are limited in their ability to address complex weather conditions due to  poor prior generalization. 
Recently, convolutional neural network (CNN) based methods have been proposed to handle the problems of image deraining~\cite{fu2017clearing,wang2019spatial,you2015adherent}, dehazing~\cite{cai2016dehazenet,ren2016single,zhang2018densely}, and desnowing~\cite{liu2018desnownet,li2019stacked,zhang2021deep}. These methods focus on learning a mapping from the weather-degraded image to the restored image using specific architectural designs, such as residual learning~\cite{liu2019dual,jiang2020multi}, multi-scale or multi-stage networks~\cite{dong2020multi,zhang2021deep}, dense connections~\cite{liu2019griddehazenet,zhang2021deep}, GAN structure~\cite{qu2019enhanced,jaw2020desnowgan}, and attention mechanism~\cite{zhang2020pyramid,zamir2021multi}. 
However, these methods are often designed for a single specific task and may not work well for multi-weather restoration. 
\begin{figure}[t]
	\centering
	 \includegraphics[width=0.45\textwidth]{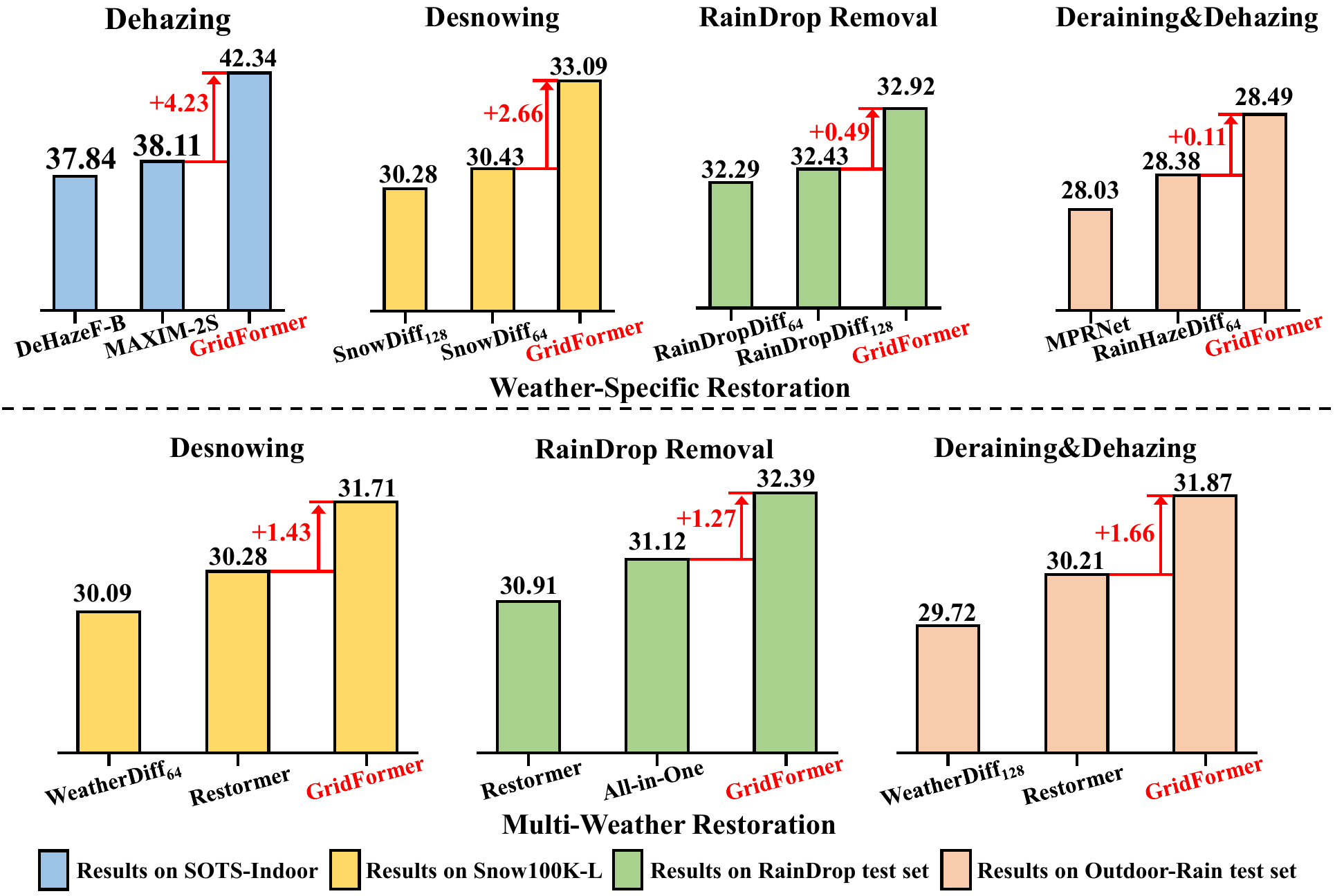}
 \caption{{\bf Comparison results for image restoration in adverse weather conditions}. Results on {\bf (top)} weather-specific restoration, and {\bf (bottom)} multi-weather restoration tasks, showing state-of-the-art performance in terms of PSNR.
 }
	\label{fig:performance}
\end{figure}

Recently, a new approach has emerged to address the challenge of multi-weather restoration in a unified architecture~\cite{li2020all,valanarasu2022transweather,li2022all,ozdenizci2023restoring}. The pioneering work of Li \etal \cite{li2020all} proposes a multi-encoder and decoder network, with each encoder dedicated to processing one type of degradation. The network is optimized using neural architecture search.
Subsequent works have borrowed this structure to improve multi-weather restoration performance. For instance, Valanarasu \etal \cite{valanarasu2022transweather} introduced the TransWeather network that employs self-attention for multi-weather restoration. Although TransWeather is more efficient than the task-specific encoder network, its performance is constrained by its inadequate exploitation of feature fusion across different scales in the network. Recently, some works focus on designing the general backbone network to exploit multi-scale features in the network for vision tasks. For example, HRNet~\cite{wang2020deep} and HRFormer~\cite{NEURIPS2021_3bbfdde8} are built by multi-resolution parallel design to learn high-resolution representations. RevCo~\cite{cai2022reversible} adopts the design of using columns (each column is a subnetwork), which aims to learn disentangled representations. These methods work well on human pose estimation, semantic segmentation, object detection, \textit{etc.} However, there are currently no specifically designed transformer-based methods to effectively utilize these features to recover degraded images under severe weather conditions.


In this paper, we propose GridFormer, a transformer-based network for image restoration in adverse weather conditions. 
GridFormer uses residual dense transformer blocks (RDTB) embedded in a grid structure to exploit hierarchical image features. The RDTB, as the key unit of the GridFormer, contains compact-enhanced transformer layers with dense connections, and local feature fusion with local skip connections.
The compact-enhanced transformer layer employs a sampler and compact self-attention for efficiency and a local enhancement stage for strengthening local details. We evaluate GridFormer on weather degradation benchmarks, including RainDrop~\cite{qian2018attentive}, SOTS-indoor~\cite{li2018benchmarking}, Haze4K~\cite{liu2021synthetic}, Outdoor-Rain~\cite{li2019heavy}, and Snow100K~\cite{liu2018desnownet}, see Fig.~\ref{fig:performance}.





In summary, the contributions of this work are three-fold:
\begin{itemize}
  \item \textbf{Unified Framework}:
    We propose a novel and unified framework called GridFormer, which is tailored specifically for image restoration under adverse weather conditions. This innovative framework seamlessly integrates residual dense transformer blocks (RDTBs) with a grid structure, creating a comprehensive architecture. Notably, incorporating RDTBs within a grid structure enables GridFormer to capture hierarchical image features efficiently. The grid structure facilitates the integration of contextual information from various spatial scales, enhancing the network's ability to restore images effectively. 

  \item \textbf{Compact-enhanced Self-Attention}: GridFormer introduces the compact-enhanced self-attention mechanism, a critical contribution. This mechanism enhances the local modeling capacity of transformer units, enabling GridFormer to capture fine-grained details in adverse weather conditions while improving network efficiency.

\item \textbf{State-of-the-art Performance}: We show the general applicability of our GridFormer by applying it to five diverse image restoration tasks in adverse weather conditions, including image deraining, image dehazing, image deraining $\&$ dehazing, desnowing, and multi-weather restoration. Our GridFormer achieves a new state-of-the-art on both weather-specific and multi-weather restoration tasks. 
\end{itemize} 

The remainder of this paper is organized as follows: Sec.~\ref{sec:related_work} discusses the related work. Sec.~\ref{sec:method} introduces our proposed method. Then, experimental results are reported and analyzed in Sec.~\ref{sec:experiment}. Sec.~\ref{sec:limitations} discusses limitations and future work. Finally, Sec.~\ref{sec:conclusion} presents a conclusion of this paper.

\section{Related Work}\label{sec:related_work}
The proposed method is related to image restoration in adverse weather conditions and transformer architecture, which are reviewed in the following.
\subsection{Restoration in Adverse Weather Conditions}
Image restoration in adverse weather conditions is the task of restoring a high-quality image under weather-related foreground degradations like rain, fog, and snow. Especially, image restoration in adverse weather conditions typically includes image deraining~\cite{ba2022not,kang2011automatic,li2019heavy,yang2019joint,wang2020model}, image dehazing~\cite{berman2016non,cai2016dehazenet,ren2018gated,liu2019griddehazenet}, image desnowing~\cite{ren2017video,liu2018desnownet,zhang2021deep}, and multi-weather restoration~\cite{li2020all,valanarasu2022transweather,ozdenizci2023restoring}. The traditional model-based methods~\cite{he2010single,luo2015removing,zhu2017joint} focus on exploring appropriate weather-related priors to address the image restoration problem. However, there has been a surge in the number of data-driven methods proposed in recent years. Next, we mainly discuss these data-driven methods in detail.    

\textbf{Deraining:}  
The task of removing rain streaks from images has been approached using a deep network called DerainNet, proposed by Fu \etal \cite{fu2017clearing}. This approach learns the nonlinear mapping between clean and rainy detail layers. Several techniques have been proposed to improve performance, such as the recurrent context aggregation in RESCAN~\cite{li2018recurrent}, spatial attention in SPANet~\cite{wang2019spatial}, multi-stream dense architecture in DID-MDN~\cite{zhang2018density}, conditional GAN-based method in~\cite{zhang2019image}, and conditional variational deraining based on VAEs~\cite{du2020conditional}. Another approach to image deraining is removing raindrops. Yamashita \etal \cite{yamashita2005removal} developed a stereo system to detect and remove raindrops, while You \etal \cite{you2015adherent} proposed a motion-based method. Qian \etal \cite{qian2018attentive} developed a raindrop removal benchmark and proposed an attentive GAN. Quan \etal \cite{quan2019deep} introduced an image-to-image CNN embedded attention mechanism to recover rain-free images, and Liu \etal \cite{liu2019dual} designed a dual residual network to remove raindrops. Zhang \etal \cite{zhang2021multifocal} proposed a multifocal attention-based cross-scale network that employs spatial and channel attention to explore cross-scale correlations of rain streaks and background for image draining. Recent works aim to remove both streaks and raindrops from images simultaneously~\cite{quan2021removing,xiao2022image}.

\textbf{Dehazing:} 
Two pioneering methods for image dehazing are DehazeNet~\cite{cai2016dehazenet} and MSCNN~\cite{ren2016single}, which first 
estimate the transmission map and generate haze-free images using an atmosphere scattering model~\cite{narasimhan2000chromatic}. AOD-Net~\cite{li2017aod} represents another advancement, which estimates one variable from the transmission map and atmospheric light. DCPDN~\cite{zhang2018densely} employs two sub-networks to estimate the transmission map and the atmospheric light, respectively. Recent works have focused on directly restoring clear images from hazy images, using attention mechanisms~\cite{qin2020ffa,zhang2020pyramid}, multi-scale structures~\cite{dong2020multi,liu2019griddehazenet}, GAN structures~\cite{qu2019enhanced} and transformers~\cite{song2022vision}. 
The network in~\cite{liu2019griddehazenet} is a similar method to our GridFormer. However, GridFormer significantly differs from~\cite{liu2019griddehazenet} in several ways. First, GridFormer is the first transformer-based method for image restoration in adverse weather conditions, whereas~\cite{liu2019griddehazenet} is a CNN-based method specifically designed for image dehazing. GridFormer is more general in terms of its utility. Second, in each GridFormer layer, we design a novel compact-enhanced transformer layer and integrate it in a residual dense manner. This promotes feature reuse and consequently enhances feature representation, whereas~\cite{liu2019griddehazenet} uses existing residual dense blocks in its network. Finally, extensive experiments demonstrate the superior performance of GridFormer compared to the method in~\cite{liu2019griddehazenet}.

\textbf{Desnowing:} In DesnowNet~\cite{liu2018desnownet}, translucency and residual generation modules were employed to restore image details. Li \etal \cite{li2019stacked} proposed a stacked dense network with a multi-scale structure.
Chen \etal \cite{chen2020jstasr} introduced a desnowing method called JSTASR, which is specifically developed for size- and transparency-aware snow removal. They used a joint scale and transparency-aware adversarial loss to improve the quality of the desnowed images.
Li \etal \cite{li2020all} adopted the network architecture search technique to obtain excellent results. 
Zhang \etal \cite{zhang2021deep} proposed a dense multi-scale desnowing network that incorporates learned semantic and geometric priors.
More recently, some works~\cite{chen2022snowformer,zhang2022desnowformer} have explored the transformer architecture and further improved the performance.

\textbf{Multi-weather restoration:} 
Beyond the above task-specific image restoration methods, recent works~\cite{li2020all,valanarasu2022transweather,li2022all} attempt to address multi-weather restoration in a single architecture. Li \etal \cite{li2020all} proposed All-in-One networks with a multi-encoder and decoder structure to restore adverse multi-weather degraded images. Specifically, they adopt separate encoders for different weather degradations and resort to neural architecture search to seek the best task-specific encoder. In \cite{li2022all}, All-in-one restoration network consists of a contrastive degraded encoder and a degradation-guided restoration network. Valanarasu \etal \cite{valanarasu2022transweather} proposed an end-to-end multi-weather image restoration model named TransWeather that achieves high performance on multi-weather restoration. The core insights in TransWeather are the intra-path transformer block and transformer decoder with learnable weather-type embeddings. In this paper, our work aligns with this direction and focuses on designing a general model to address the multi-weather restoration problem. In addition, there are methods aimed at designing effective network architecture for image restoration. For example, MPRNet~\cite{zamir2021multi} and MAXIM~\cite{tu2022maxim} are general image restoration methods that have also been successful in addressing a range of adverse weather conditions. MIMOUNet~\cite{cho2021rethinking} adopts an encoder-decoder-based U-shaped network with multi-input and multi-output to achieve image deblurring. In our method, we employ the coarse-to-fine strategy to the transformer network in the grid structure for image restoration under adverse weather conditions.

\begin{figure*}[t]
\begin{center}
	\includegraphics[width=0.99\textwidth]{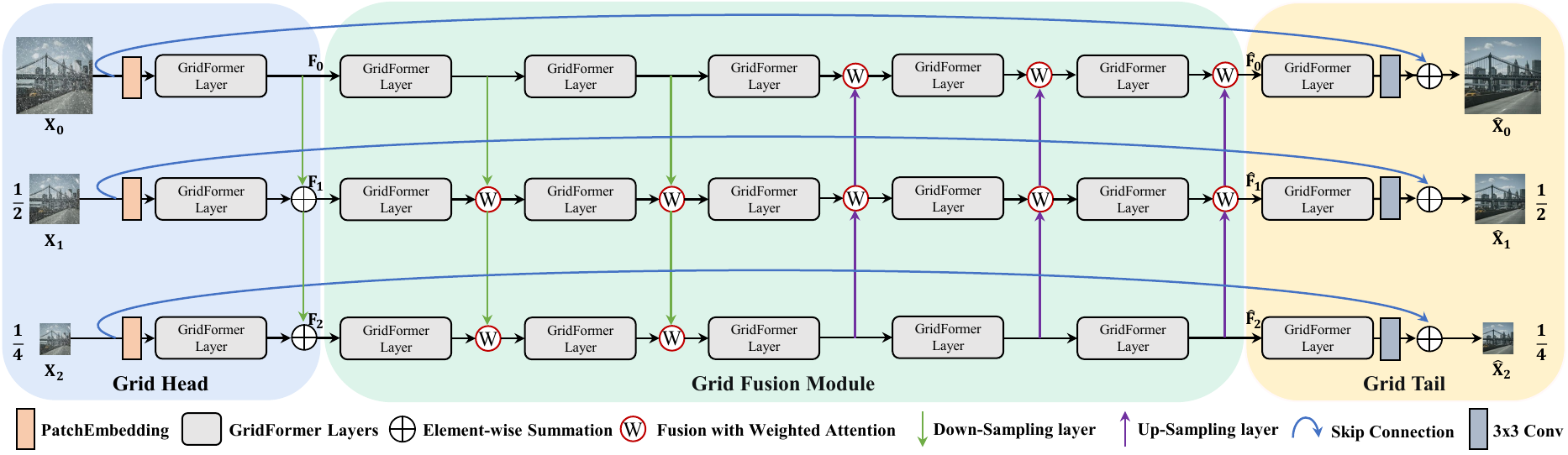}
	\caption{\textbf{GridFormer architecture}. It consists of a grid head, a grid fusion module, and a grid tail. The pyramid degraded images $\mathbf{X}_{0}, \mathbf{X}_{1}, \mathbf{X}_{2}$ are first fed into the grid head to extract hierarchical initial features $\mathbf{F}_{0}, \mathbf{F}_{1}, \mathbf{F}_{2}$. The initial features are further refined by the grid fusion module to generate features $\hat{\mathbf{F}}_{0}, \hat{\mathbf{F}}_{1}, \hat{\mathbf{F}}_{2}$. Finally, the gird tail reconstructs clear images $\hat{\mathbf{X}}_{0}, \hat{\mathbf{X}}_{1}, \hat{\mathbf{X}}_{2}$.}
	\label{fig:overall}
 \end{center}
  \vspace{-0.3cm}
\end{figure*}
\subsection{Vision Transformers in Image Restoration}
Recently, vision transformers have witnessed great success in low-level image restoration. Specifically, inspired by the seminal work in~\cite{vaswani2017attention}, Chen \etal \cite{chen2021pre} proposed an Image Processing Transformer (IPT) for general image restoration, which employs a special multi-head and multi-tail structure to adapt for the specific image restoration tasks. However, IPT requires costly pre-training on large-scale datasets. Further, SwinIR~\cite{liang2021swinir} and Uformer~\cite{wang2022uformer} modify the original Swin Transformer block and obtain good performance with relatively low computational cost. In particular, SwinIR stacks the proposed residual transformer blocks to extract deep features for image reconstruction. Uformer adopts a U-shape structure, embedding the proposed LeWin transformer blocks to predict residual images. Yao \etal \cite{yao2022dense} adopted the LeWin transformer block as a basic unit and introduced the dense residual skip connection to propose a dense residual skip-connection network based on transformer called DenSformer for image denoising. Liang \etal \cite{liang2022drt} proposed a recursive transformer, which first introduces a recursive local window-based self-attention structure in the network. A recent method, Restormer~\cite{zamir2021restormer}, which is a multi-scale hierarchical transformer architecture, has also yielded fine restoration performance on image restoration \eg deraining. Inspired by the success of these methods, we propose a general grid framework with novel transformer blocks to restore images in adverse weather conditions. SwinIR and DenSformer are similar methods to our GridFormer. However, while SwinIR fuses Swin Transformer and convolutional layers in its residual Swin Transformer block, our GridFormer's residual dense block more effectively enhances feature reuse. Unlike DenSformer's dense residual transformer block, our approach is characterized by the unique compact-enhanced self-attention mechanism, local feature fusion, and local skip connections within the residual dense transformer block.
\section{Method}\label{sec:method}
To explore the potential use of the transformer on image restoration in adverse weather conditions for obtaining better results, we propose the GridFormer by embedding residual dense transformer blocks in a grid structure. The motivation and overall architecture of the proposed GridFormer will firstly be introduced in Sec.~\ref{sec:method:architecture}, and then the core component (\ie residual dense transformer block) of our GridFormer will be discussed in Sec.~\ref{sec:method:rdtb}. Finally, the loss functions will be presented in Sec.~\ref{sec:loss_function}.

\subsection{Motivation and Architecture}\label{sec:method:architecture}
\textbf{Motivation}. Our motivation arises from the urgent need for techniques that restore images captured in unfavorable weather conditions. Weather-related factors, such as haze, rain, and snow, significantly impact the quality and perception of images, which in turn affects various practical applications such as surveillance, autonomous driving, and outdoor photography. The main objective of developing the proposed GridFormer is to address the persistent challenges caused by adverse weather conditions on image quality. Our goal is to create an image restoration framework that effectively handles a range of adverse weather scenarios, thereby enhancing the quality of images affected by these conditions.

\textbf{Architecture}. As shown in Fig.~\ref{fig:overall},  GridFormer contains three paths from the weather-degraded images to the recovered ones, where each path conducts restoration at different image resolutions. In GridFormer, the higher resolution path continuously interacts dynamically with the lower resolution path in the network to remove weather degradation accurately, and the lower resolution path provides useful global information owing to larger receptive fields. Each path is composed of seven GridFormer layers. Different paths are interlinked with a down-sampling layer, an up-sampling layer, and weighted attention fusion units to compose the columns of the GridFormer. Thanks to the grid structure with three rows and seven columns, information from different resolutions can be shared effectively. Specifically, GridFormer consists of three parts: grid head (GH), grid fusion module (GFM), and grid tail (GT). We present the details of each part in the following.

\textit{Grid head.} To extract initial multi-resolution features, we use a grid head architecture to process pyramid input images in parallel. Every path in the grid head consists of a feature embedding layer, achieved by $3\times3$ convolutions, and a GridFormer layer. As shown in Fig.~\ref{fig:overall}, given a weather-degraded image $\mathbf{X}_{0}$, the grid head extracts hierarchical features $\mathbf{F}=\left\{\mathbf{F}_{0}, \mathbf{F}_{1}, \mathbf{F}_{2} \right\}$ in different channels (\ie $C$, $2C$, and $4C$) from pyramid images $\mathbf{X}=\left\{\mathbf{X}_{0}, \mathbf{X}_{1}, \mathbf{X}_{2} \right\}$ (1/2, 1/4 scales for $\mathbf{X}_{1}$ and $\mathbf{X}_{2}$). In our experiments, we use $C=48$. The grid head computation can be defined as:
\begin{equation}
\small
\setlength\abovedisplayskip{8pt}%
\setlength\belowdisplayskip{8pt}
\mathbf{F}_{i}= \begin{cases}\mathrm{GFL}_{i}\left(\mathrm{E}_{i}(\mathbf{X}_{0}) \right), & i=0 \\ \mathrm{GFL}_{i}(\mathrm{E}_{i}(\mathbf{X}_{i}))+(\mathbf{F}_{i-1})_{\downarrow}, & i=1,2 \end{cases}
\end{equation}
where $i$ is the $i$-th network path, and $\mathrm{E}_{i}$ is the feature embedding layer. The $\downarrow$ symbol denotes the down-sampling layer, where we use a $3\times3$ convolution with a pixel-unshuffle operation~\cite{shi2016real} to halve the features in the spatial dimensions while doubling the channels. $\mathrm{GFL}$ is a GridFormer layer that is mainly built from residual dense transformer blocks. 

\begin{figure*}[t]
\begin{center}
	\includegraphics[width=0.98\textwidth]{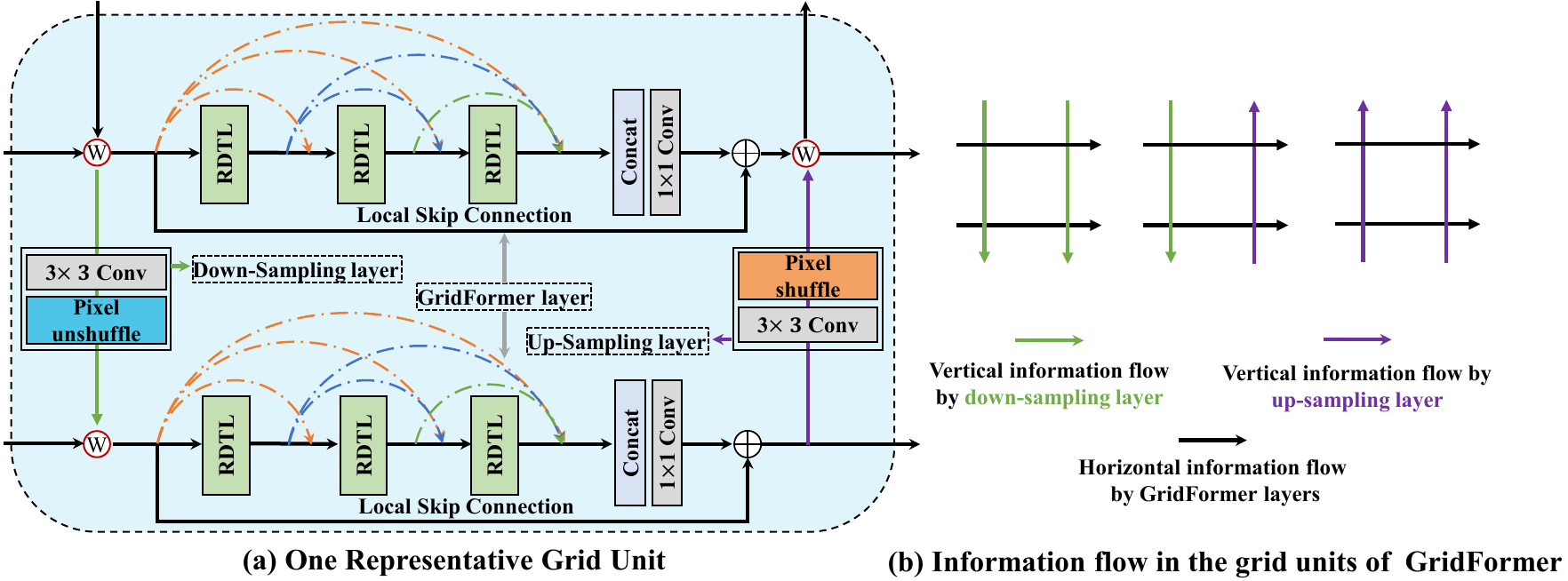}
	\caption{\textbf{Grid unit structure and information flow}. (a) The structure of a single grid unit is comprised of four parts: the down-sampling layer, the GridFormer layer, the up-sampling layer, and attention fusion operations. RDTL refers to the proposed residual dense transformer layer. (b) Information flow of grid units in the fusion module.}
	\label{fig:grid_unit}
 \end{center}
\end{figure*}

\textit{Grid fusion module}. To fully integrate the hierarchical features of different rows and columns in the network, we propose a grid fusion module between the grid head and the grid tail. The structure of the proposed grid fusion module is organized into a 2D grid pattern. As illustrated in Fig.~\ref{fig:overall}, the fusion module is designed in a grid-like structure of three rows and five columns. In particular, each row contains five consecutive GridFormer layers that keep the feature dimension constant. In the column axis, according to the position in the grid, we resort to the down-sampling layers or up-sampling layers to change the size of the feature maps for feature fusion. Fig.~\ref{fig:grid_unit}~(a) shows a representative grid unit in the fusion module. The GridFormer layer is a dense structure consisting of three residual dense transformer layers (RDTL) and a $1\times1$ convolution, which will be discussed in the next subsection. The down-sampling and up-sampling layers are symmetrical and use a $3\times3$ convolution with pixel-shuffle or pixel-unshuffle operation~\cite{shi2016real} to change the feature dimensions. In addition, considering that the features of different scales may not be equally important, we use a simple weighted attention fusion strategy to achieve feature fusion from the different row and column dimensions. Inspired by~\cite{zheng2022t,wang2022ultra}, we first generate two trainable weights for different features, where each parameter is an $n$-dimensional vector ($n$ is the channels of feature). We add these weighted features to derive the fusion features. Grid units in the grid fusion module provide different information flows for feature fusion shown in Fig.~\ref{fig:grid_unit}~(b), which guides the network to produce better-recovered results in combination with different complementary information.
\begin{figure}[t]
\begin{center}
	\includegraphics[width=0.48\textwidth]{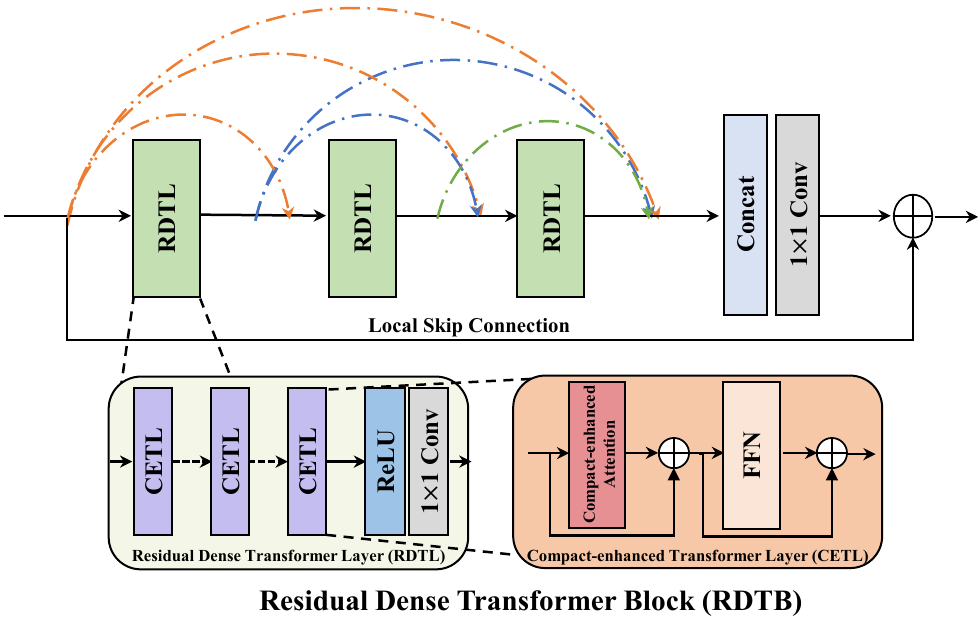}
	\caption{\textbf{The structure of the proposed Residual Dense Transformer Block (RDTB)}. It includes three residual dense transformer layers, a $1\times1$ convolution for local feature fusion, and a local skip connection for local residual learning. The residual dense transformer layer is mainly built by the proposed compact-enhance transformer layer, which contains the compact-enhanced self-attention and FFN.}
	\label{fig:rdtb}
 \end{center}
\end{figure}
\begin{figure*}[t]
\begin{center}
	\includegraphics[width=\textwidth]{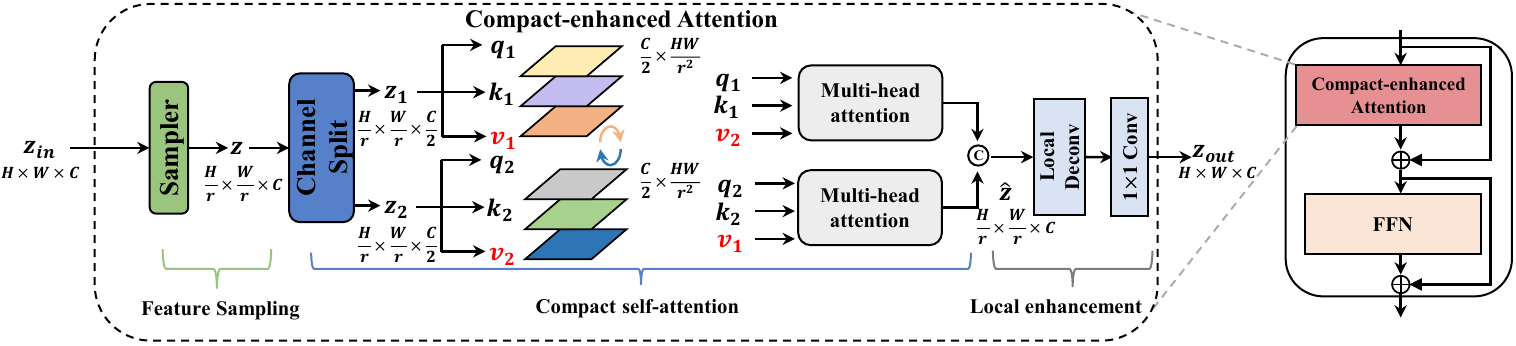}
	\caption{\textbf{Right}: the schematic illustration of the proposed Compact-enhanced Transformer Layer consisting of a compact-enhanced attention and a Feed-Forward Network (FFN). \textbf{Left}: the compact-enhanced attention layer, which contains three steps, feature sampling, compact self-attention, and local enhancement. $H$, $W$, and $C$ denote the height, width, and numbers of feature channels, respectively. $r$ is the feature sampling rate. $\copyright$ and $\oplus$ refer to concatenate and element-wise summation operations respectively. }
	\label{fig:cetl}
 \end{center}
 \vspace{-0.3cm}
\end{figure*}

\textit{Grid tail}. To further improve the quality of the recovered images, we design a grid tail module to predict multi-scale outputs.  The structure of the grid tail is symmetrical to that of the grid head. Specifically, each path is composed of a GridFormer layer, a $3\times3$ convolution, and a long skip connection for image reconstruction. The skip connection is used to transmit input information directly to the grid tail module, which maintains the color and detail of the original image. The complete process is formulated as:
\begin{equation}
\small
\hat{\mathbf{X}}_{i}=\mathrm{C}_{i}(\mathrm{GFL}_{i}(\hat{\mathbf{F}}_{i}))+\mathbf{X}_{i}, i \in\{0,1,2\},
\end{equation}
where $\hat{\mathbf{X}}_{i}$ is the final result of GridFormer on the $i$-th path, $\mathrm{C}_{i}$ is a $3\times3$ convolution, and $\hat{\mathbf{F}}_{i}, i \in\{0,1,2\}$ is the output feature of the grid fusion module. To optimize the network parameters, we train GridFormer using a combination of two losses, multi-scale Charbonnier loss~\cite{charbonnier1994two} and perceptual loss, where the weight of perceptual loss~\cite{johnson2016perceptual} is set to $0.1$. Next, we detail the core component residual dense transformer block that is used to build the elemental layer of GridFormer.  

\subsection{Residual Dense Transformer Block}\label{sec:method:rdtb}
Previous works~\cite{huang2017densely,zhang2018residual,liu2019griddehazenet,zhang2021deep,zheng2022t} have shown that using dense connections has many advantages,  mitigating the vanishing gradient problem, encouraging feature reuse and enhancing information propagation. Accordingly, we propose to design the transformer with dense connections to build the basic GridFormer layers. Specifically, we propose residual dense transformer blocks (RDTB) to compose GridFormer using different settings. As illustrated in Fig.~\ref{fig:rdtb},  RDTB contains densely connected transformer layers, local feature fusion, and local residual learning. When implementing the dense connection, we mainly incorporate three layers of residual dense transformer layers (RDTL), with the growth rate set at 16. This implies that each individual RDTL generates 16 new feature maps. These newly generated feature maps are subsequently concatenated with the feature maps received from the preceding layer. Within each RDTL, we use several compact-enhanced transformer layers (CETL) with a ReLU~\cite{glorot2011deep} activation function to extract features, and adopt a $1\times1$ convolution to ensure the same number of channels for input and output features. For local feature fusion and local residual learning, we introduce a $1\times1$ convolution and a local skip connection in RDTB to control the final output.

The direct application of transformers~\cite{vaswani2017attention,dosovitskiyimage} to our grid network will lead to high computational overhead, we thus develop a cost-effective compact-enhanced attention, with the stages of sampler and compact self-attention for improving the efficiency, as well as a local enhancement stage for enhancing the local information in the transformer. Fig.~\ref{fig:cetl} illustrates the detailed structure of the proposed compact-enhanced attention.

\textbf{Feature sampling}. We first design a sampler to produce down-sampled input tokens for the subsequent self-attention computation. The sampler is built by an average pooling layer with stride $r$. The sampler layer not only increases the receptive field to observe more information, but also enhances the invariance on the input token. In addition, the produced lower-resolution features can reduce the computation of subsequent layers. The feature sampling step is formulated as:
\begin{equation}
\label{eq3}
\small
\setlength\abovedisplayskip{8pt}%
\setlength\belowdisplayskip{8pt}
\mathbf{Z}=\mathrm{Avg}_{r}(\mathbf{Z}_{in}),
\end{equation}
where $\mathbf{Z}_{in} \in \mathbb{R}^{H \times W \times C}$ represents the input token. $\mathbf{Z} \in \mathbb{R}^{\frac{H}{r} \times \frac{W}{r} \times C}$ is the output token. $\mathrm{Avg}_{r}$ indicates the average pooling operation with stride $r$. In the experiments, we empirically set $r$ as $4$, $2$, and $2$ in three rows of GridFormer layers, respectively (see Sec.~\ref{ablation_study}).

\textbf{Compact self-attention}. Given a feature of dimensions $H \times W \times C$, recent low-level transformer-based methods~\cite{valanarasu2022transweather,lee2022knn,wang2022uformer} aim to explore the long-range dependence between key and query to calculate the $N\times N$
attention map ($N= H\times W$), which leads to high complexity and fails to model the global information from the channel dimension. Thus, for more efficient computation in self-attention, we resort to a different strategy. Specifically, as illustrated in Fig.~\ref{fig:cetl}, for an output feature $\mathbf{Z} \in \mathbb{R}^{\frac{H}{r} \times \frac{W}{r} \times C}$ from the sampler, we first implement the split operation by dividing it along the channel dimension to produce $\mathbf{z}_{1}\in \mathbb{R}^{\frac{H}{r} \times \frac{W}{r} \times \frac{C}{2}}$ and $\mathbf{z}_{2}\in \mathbb{R}^{\frac{H}{r} \times \frac{W}{r} \times \frac{C}{2}}$. We then apply a convolution layer with reshape operation on $\mathbf{z}_{1}$ and $\mathbf{z}_{2}$, which projects $\mathbf{z}_{1}$ and $\mathbf{z}_{2}$ into Queries ($\mathbf{q}_{1},\mathbf{q}_{2} \in \mathbb{R}^{ \frac{C}{2}\times \frac{HW}{r^{2}} }$), Keys ($\mathbf{k}_{1},\mathbf{k}_{2}\in \mathbb{R}^{ \frac{C}{2}\times \frac{HW}{r^{2}}}$) and Values $\mathbf{v}_{1},\mathbf{v}_{2}\in \mathbb{R}^{ \frac{C}{2}\times \frac{HW}{r^{2}}}$), respectively. Inspired by existing methods \cite{petit2021u,susladkar2023gafnet,gu2023adafuse,zhang2024cf}, we exchange the values produced by them to perform multi-head self-attention, which can improve the interaction between $\mathbf{z}_{1}$ and $\mathbf{z}_{2}$. Compared with the method of exchanging queries for feature interaction in cross-attention~\cite{petit2021u,zhang2024cf}, our approach exchanges the values for interaction and feature fusion, finding it beneficial for better restoration performance (see Sec.~\ref{ablation_study}). Finally, we obtain the result $\hat{\mathbf{Z}}$ by concatenating the output of the two multi-head self-attention and changing their dimensions. The proposed compact self-attention mechanism can be formulated as:
\begin{equation}
\small
\setlength\abovedisplayskip{6pt}%
\setlength\belowdisplayskip{6pt}
\!\hat{\mathbf{Z}}=[\operatorname{softmax}_1\left(\frac{\mathbf{q}_{1} \mathbf{k}_{1}^{\top}}{\sqrt{d_{k_{1}}}}\right)\mathbf{v}_{2},\!\operatorname{softmax}_2\left(\frac{\mathbf{q}_{2} \mathbf{k}_{2}^{\top}}{\sqrt{d_{k_{2}}}}\right)\mathbf{v}_{1}], 
\end{equation}
where $[\cdot]$ indicates the concatenation operation. The major computational overhead in transformers mainly arises from the self-attention (SA) layer. In contrast to recent transformer-based methods that employ spatial modeling for SA, the complexity of the key-query dot-product interaction grows quadratically with the spatial resolution of input, \ie $O(N\times N)$. Our proposed compact self-attention addresses this by performing SA across channels instead of the spatial dimension, resulting in cross-covariance computation across channels to produce an attention map that implicitly encodes the global context. Consequently, our compact self-attention generates an attention map of size $\mathbb{R}^{C \times C}$, instead of the huge regular attention map of size $\mathbb{R}^{N \times N}$. Thus, our compact self-attention successfully reduces complexity.

\textbf{Local enhancement}. As shown in Fig.~\ref{fig:cetl}, we add a local feature enhancement stage in the tail of compact self-attention. This stage consists of  a deconvolution operation, sometimes referred to as a ``transposed convolution," with a deconvolution for local feature propagation and a $1\times1$ convolution for local fusion:
\begin{equation}
\small
\setlength\abovedisplayskip{8pt}%
\setlength\belowdisplayskip{8pt}
\mathbf{Z}_{out} =\mathrm{Conv_{1\times1}}(\mathrm{Deconv}(\hat{\mathbf{Z}})),
\end{equation}
where $\mathbf{Z}_{out}$ is the final output. $\mathrm{Conv_{1\times1}}$ and $\mathrm{Deconv}$ are $1\times1$ convolution and deconvolution layers respectively.

\subsection{Loss Function}\label{sec:loss_function}
Inspired by existing works~\cite{yin2023multiscale,valanarasu2022transweather,ye2022perceiving,jiang2020decomposition,li2022two,yu2022frequency,ali2023vision,hsu2023wavelet,qiao2023dual}, we use a loss function combining the Charbonnier loss~\cite{charbonnier1994two} and the perceptual loss~\cite{wang2018esrgan} to train our GridFormer. We regard the Charbonnier loss as a pixel-wise loss, which is used between the recovered images and the ground truth images at each scale, and the perceptual loss is used to help our model produce visually pleasing results. The Charbonnier loss is defined as:
\begin{equation}
\mathcal{L}_{\text{char}}=\frac{1}{3}\sum_{k=0}^{2}\sqrt{\|\hat{\mathbf{X}}_{k}-\mathbf{I}_{k}\|^{2}+\varepsilon^{2}},
\end{equation}
where $\hat{\mathbf{X}}_{k}$ and $\mathbf{I}_{k}$ refer to the restored image and ground-truth image respectively, and $k$ represents the index of the image scale level in our GridFormer. The constant $\varepsilon$ is empirically set to $10^{-3}$. For the perceptual loss, following previous work~\cite{wang2018esrgan}, we adopt a pre-trained VGG19~\cite{simonyan2014very} to extract the perceptual features from the $Conv5\_4$ layer of VGG19, and then use the $L_{1}$ loss function to compute the difference between the perceptual features of the restored images and their corresponding ground truths. This effective perceptual loss focuses on capturing high-level semantic information, resulting in sharper edges and visually appealing outcomes, all while ensuring computational efficiency~\cite{wang2018esrgan}. Specifically, the perceptual loss is as follows:
\begin{equation}\label{equation_10}
	\begin{aligned}
	\mathcal{L}_{per} = \frac{1}{3}\sum_{k=0}^{2} \frac{1}{CHW} \| \phi(\hat{\mathbf{X}}_{k})-\phi (\mathbf{I}_{k}) \|_{1},
	\end{aligned}
	\end{equation}
where $C$, $H$, and $W$ denote the dimensions of the feature map obtained from the $Conv5\_4$ layer of the pretrained VGGNet $\phi$. 

The final loss function $\mathcal{L}$ to train our proposed GridFormer is shown as follows:
\begin{equation}
\mathcal{L}=\mathcal{L}_{\text{char}}+ \alpha \mathcal{L}_{\text {per}},
\end{equation}
where $\mathcal{L}_{\text{char}}$ denotes the Charbonnier loss, $\mathcal{L}_{\text{per}}$ is the perceptual loss. $\alpha$ is a hyper-parameter that is used to balance these two losses. In our experiments, it is empirically set to $0.1$.

\subsection{Differences from Existing Methods}\label{sec:existing_methods}

While HRNet~\cite{wang2020deep}, HRFormer~\cite{NEURIPS2021_3bbfdde8}, and RevCol~\cite{cai2022reversible} utilize a grid-like structure, they diverge from our GridFormer. First,  GridFormer captures multi-scale features directly from the pixel level, in contrast to HRNet and HRFormer which perform multi-scale feature extraction at the feature layer level, and RevCol, which does not incorporate a multi-scale mechanism. Second, GridFormer integrates a new self-attention mechanism to enhance the fusion of multi-scale features more effectively. This approach sets it apart from HRNet, HRFormer, and RevCol, which do not employ compact self-attention in their feature fusion processes. Third, our network is intricately designed for image restoration under adverse weather conditions, striving to produce images of superior quality. Unlike HRNet, HRFormer, and RevCol, which are not specifically engineered for this challenge, our network architecture is uniquely suited to tackle the complexities inherent in this task.

\section{Experiments and Analysis}\label{sec:experiment}
We evaluate our GridFormer for several image restoration tasks in severe weather conditions, including \textbf{(1)} image dehazing, \textbf{(2)} image desnowing, \textbf{(3)} raindrop removal, \textbf{(4)} image deraining and dehazing, and \textbf{(5)} multi-weather restoration. Specifically, in this section, we first introduce datasets, the implementation details of our GridFormer, and the comparison methods. Then, we show the restoration results of our GridFormer and the comparison with the state-of-the-art methods. Finally, we conduct extensive ablation studies to verify the effectiveness of modules in our GridFormer.
\subsection{Experimental Setup}
We evaluate GridFormer on several image restoration tasks under severe weather conditions.

\textbf{Datasets.} For image dehazing, the first setting uses ITS~\cite{li2018benchmarking} to train the model and test it on indoor SOTS~\cite{li2018benchmarking}. Another setting is training and testing on Haze4K~\cite{liu2021synthetic} covering both indoor and outdoor scenes.
Desnowing is evaluated on Snow100K~\cite{liu2018desnownet}.  RainDrop~\cite{qian2018attentive} is used for raindrop removal, and  Outdoor-Rain~\cite{li2019heavy} is used for image deraining and dehazing. For multi-weather restoration, we train the model on a combination of images degraded in adverse weather conditions similar to~\cite{ozdenizci2023restoring}. Table~\ref{tab:datasets} lists the datasets used for the different tasks. In the following, we introduce the dataset and experimental details for specific tasks for image restoration in adverse weather conditions.

\textbf{Image dehazing}. Following~\cite{liu2019griddehazenet,qin2020ffa,song2022vision,tu2022maxim}, we conduct our experiments on RESIDE~\cite{li2018benchmarking} and Haze4K~\cite{liu2021synthetic} datasets. Specifically, for the RESIDE dataset, we adopt Indoor Training Set (ITS) to train the model and test the model on the indoor set of the SOTS dataset. ITS contains $13,990$ indoor pair images and the indoor set of the SOTS dataset includes $500$ indoor pair images. For the Haze4K dataset, we follow the previous work~\cite{ye2021perceiving}. The Haze4K dataset contains $3,000$ haze and haze-free image pairs for training and $1,000$ for testing. The Haze4K dataset is more challenging, which considers both indoor and outdoor scenes.

\begin{table}[!t]
\centering
\scriptsize
\setlength{\tabcolsep}{2pt}
\caption{Dataset summary on five tasks of image restoration in adverse weather conditions.}
\scalebox{0.90}{\begin{tabular}{l|l|rr}
 \toprule
 \textbf{Task}  & \textbf{Dataset}  & \textbf{\#Train} & \textbf{\#Test} \\
 \toprule
 \multirow{3}{*}{\footnotesize\textbf{Image Dehazing}} & ITS~\cite{li2018benchmarking} & $13,990$ & $0$ \\
  & SOTS-Indoor~\cite{li2018benchmarking} & $0$ & $500$  \\
 & Haze4K~\cite{liu2021synthetic} & $3,000$ & $1,000$  \\
 \hline
 \multirow{3}{*}{\footnotesize\textbf{Image Desnowing}} & 
 Snow100K~\cite{liu2018desnownet} & $50,000$ & $0$  \\
 & Snow100K-S~\cite{liu2018desnownet} & $0$ & $16,611$  \\
 & Snow100K-L~\cite{liu2018desnownet} & $0$ & $16.801$  \\
 \hline
 \multirow{2}{*}{\footnotesize\textbf{Raindrop Removal}}
 & RainDrop~\cite{qian2018attentive}  & $861$ & $0$  \\
  & RainDrop-Test~\cite{qian2018attentive}  & $0$ & $51$  \\
 \hline
{\footnotesize\textbf{Image Deraining \&}} 
 & Outdoor-Rain~\cite{li2019heavy}  & $9,000$& $0$   \\
{\footnotesize\textbf{Image Dehazing}}  & Outdoor-Rain-Test~\cite{li2019heavy}  & $0$ & $750$   \\
 \hline
 \multirow{5}{*}{\footnotesize\textbf{Multi-weather Restoration}} & 
All-weather~\cite{li2022all} & $18,069$ & $0$  \\
 & Snow100K-S~\cite{liu2018desnownet} & $0$ & $16,611$  \\
 & Snow100K-L~\cite{liu2018desnownet} & $0$ & $16.801$  \\
 & RainDrop-Test~\cite{qian2018attentive}  & $0$ & $51$  \\
 & Outdoor-Rain-Test~\cite{li2019heavy}  & $0$ & $750$  \\
\bottomrule
\end{tabular}}
 \vspace{-0.3cm}
\label{tab:datasets}
\end{table}
\textbf{Image desnowing}. For this task, we use the popular Snow100K dataset~\cite{liu2018desnownet} for training and evaluating the proposed method. Snow100K contains $50,000$ training and $50,000$ testing images. The testing set has three sub-sets \ie Snow100K-S/M/L, which refers to different snowflake sizes (light/mid/heavy). The Snow100K-S, Snow100K-M, and Snow100K-L have $16611$, $16588$, and $16801$ image pairs, respectively. In our experiment, we keep the same setup of~\cite{ozdenizci2023restoring}. Specifically, we use the training set to train our model and evaluate the proposed method on Snow100K-S and Snow100K-L.

\textbf{Raindrop removal}. Consistent with previous works~\cite{qian2018attentive,li2022all,valanarasu2022transweather,ozdenizci2023restoring}, we adopt a representative RainDrop dataset~\cite{qian2018attentive} for raindrop removal. The RainDrop dataset includes $861$ synthetic raindrop training images and $58$ images for testing.  

\textbf{Image deraining and dehazing}. For this task, we train our GridFormer with 
Outdoor-Rain dataset~\cite{li2019heavy}, which considers dense synthetic rain streaks and provides realistic scene views. Therefore, this dataset is designed to solve the problem of image deraining and dehazing. It consists of $9,000$ images for training and $750$ for testing.

\textbf{Multi-weather restoration}. Following the previous works~\cite{li2022all,valanarasu2022transweather,ozdenizci2023restoring}, we use a mixed dataset called All-weather, in which the training set contains $18,069$ images sampled from Snow100K~\cite{liu2018desnownet}, Raindrop~\cite{qian2018attentive}, and Outdoor-Rain~\cite{li2019heavy}. We use the Snow100k-S/L test sets to evaluate the model's performance for the image desnowing task. In addition, we adopt the testing sets of the RainDrop dataset and Outdoor-Rain dataset to evaluate the model's performance for the raindrop removal task and image deraining \& dehazing task, respectively.

\begin{figure*}[t]
\begin{center}
	\includegraphics[width=\textwidth]{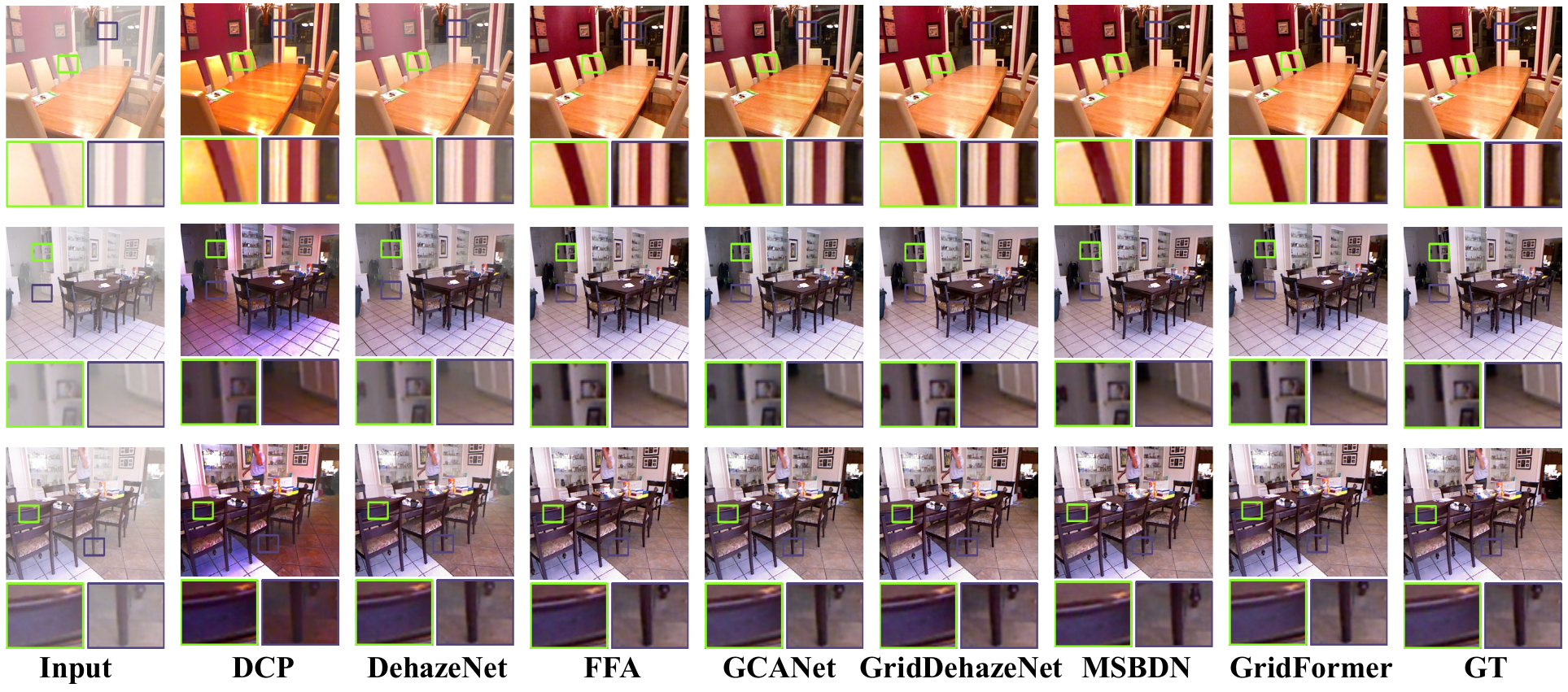}
	\caption{Dehazing comparison on SOTS-indoor. From left to right are the input images, results of DCP~\cite{he2010single}, DehazeNet~\cite{cai2016dehazenet}, FFA-Net~\cite{qin2020ffa}, GCANet~\cite{chen2019gated}, GridDehazeNet~\cite{liu2019griddehazenet}, MSBDN~\cite{dong2020multi}, our GridFormer, and ground truth images, respectively. The images restored by GridFormer are more clear and closer to the ground truth. \textbf{Zoom in for details}.}
	\label{fig:dehazing-visuals}
 \end{center}
\end{figure*}
\begin{figure*}[t]
\begin{center}
	\includegraphics[width=\textwidth]{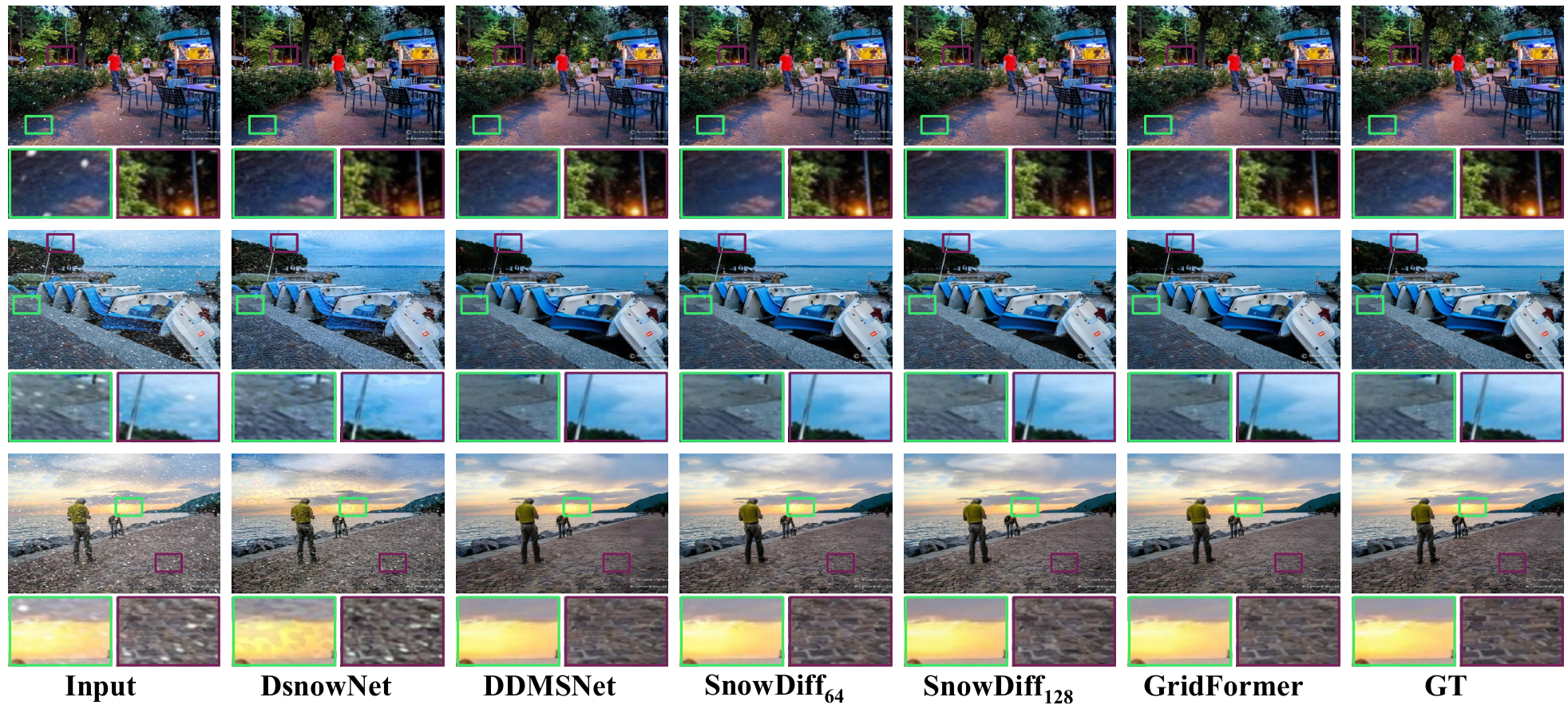}
	\caption{Desnowing comparison on Snow100K-S test set. From left to right are the input images, results of DesnowNet~\cite{liu2018desnownet}, DDMSNet~\cite{zhang2021deep}, SnowDiff$_{64}$~\cite{ozdenizci2023restoring}, SnowDiff$_{128}$~\cite{ozdenizci2023restoring}, our GridFormer, and ground truth images, respectively. \textbf{Zoom in for details}.}
	\label{fig:desnowing-visuals-s}
 \end{center}
\end{figure*}
\begin{figure*}[t]
\begin{center}
	\includegraphics[width=\textwidth]{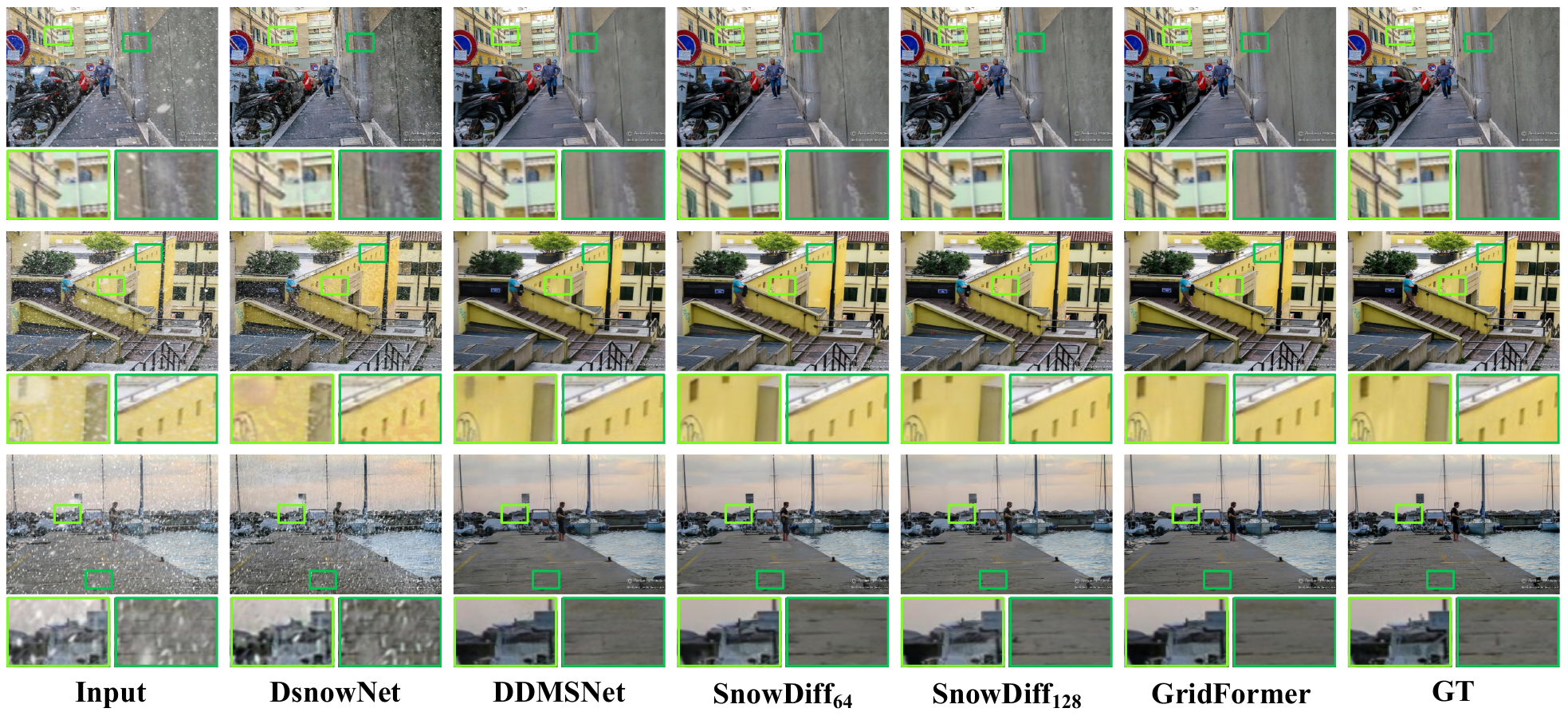}
	\caption{Desnowing comparison on Snow100K-L test set. From left to right are the input images, results of DesnowNet~\cite{liu2018desnownet}, DDMSNet~\cite{zhang2021deep}, SnowDiff$_{64}$~\cite{ozdenizci2023restoring}, SnowDiff$_{128}$~\cite{ozdenizci2023restoring}, our GridFormer, and ground truth images, respectively. \textbf{Zoom in for details}.}
	\label{fig:desnowing-visuals-l}
 \end{center}
\end{figure*}

\textbf{Implementation details.} We implemented GridFormer in PyTorch,
using the AdamW optimizer~\cite{loshchilovdecoupled} with $\beta_{1} = 0.9$ and $\beta_{2} = 0.999$. The learning rate is set to $3\times10^{-4}$ and decreased to $10^{-6}$ using the cosine annealing decay strategy~\cite{loshchilov2016sgdr}. For each task, we train the model with different iterations and patch sizes. At training time we use random horizontal and vertical flips for data augmentation. Following the setup in~\cite{xiao2022image,valanarasu2022transweather,tu2022maxim}, we evaluate the performance by PSNR and SSIM calculated in RGB space for image dehazing, and on the Y channel for other tasks.

\textbf{Comparison methods}. The comparison methods for the image dehazing task are traditional method DCP~\cite{he2010single}, CNN-based methods DehazeNet~\cite{cai2016dehazenet}, MSCNN~\cite{ren2016single}, AOD-Net~\cite{li2017aod}, GFN~\cite{ren2016single}, GCANet~\cite{chen2019gated}, GridDehazeNet~\cite{liu2019griddehazenet}, MSBDN~\cite{dong2020multi}, PFDN~\cite{dong2020physics}, FFA-Net~\cite{qin2020ffa}, and AECR-Net~\cite{wu2021contrastive}, and recent transformer-based methods DehazeF-B~\cite{song2022vision}, and MAXIM-2S~\cite{tu2022maxim}. For the task of image desnowing, the comparison methods are SPANet~\cite{wang2019spatial}, JSTASR~\cite{chen2020jstasr}, RESCAN~\cite{li2018recurrent}, DesnowNet~\cite{liu2018desnownet}, DDMSNet~\cite{zhang2021deep}, SnowDiff$_{64}$~\cite{ozdenizci2023restoring}, and SnowDiff$_{128}$~\cite{ozdenizci2023restoring}. As for the raindrop removal task, the comparison methods are pix2pix~\cite{isola2017image}, DuRN~\cite{liu2019dual}, RaindropAttn~\cite{quan2019deep}, AttentiveGAN~\cite{qian2018attentive}, IDT~\cite{xiao2022image}, RainDropDiff$_{64}$~\cite{ozdenizci2023restoring}, and RainDropDiff$_{128}$~\cite{ozdenizci2023restoring}. The comparison methods for the image deraining and dehazing task are CycleGAN~\cite{zhu2017unpaired}, pix2pix~\cite{isola2017image}, HRGAN~\cite{li2019heavy}, PCNet~\cite{jiang2021rain}, MPRNet~\cite{zamir2021multi}, RainHazeDiff$_{64}$~\cite{ozdenizci2023restoring}, and RainHazeDiff$_{128}$~\cite{ozdenizci2023restoring}. Finally, the comparison methods for the multi-weather restoration task are All-in-One \cite{li2020all}, TransWeather~\cite{valanarasu2022transweather}, Restormer~\cite{zamir2021restormer}, WeatherDiff$_{64}$~\cite{ozdenizci2023restoring}, and WeatherDiff$_{128}$~\cite{ozdenizci2023restoring}.

\begin{table}[t]
\centering
\scriptsize
\caption{Dehazing results on SOTS-indoor and Haze4K. Bold and underlined fonts denote the best and second-best results, respectively.}
\setlength{\tabcolsep}{5pt}
\renewcommand{\arraystretch}{1.1}
\scalebox{0.85}{\begin{tabular}{c||c|cc|cc}
\hline
\rowcolor{mygray}
&& \multicolumn{2}{c}{SOTS-Indoor~\cite{li2018benchmarking}} & \multicolumn{2}{c}{Haze4K~\cite{liu2021synthetic}}  \\ 
\rowcolor{mygray}
\multirow{-2}*{Type}&\multirow{-2}*{Method} & PSNR$\uparrow$ & SSIM$\uparrow$ & PSNR$\uparrow$ & SSIM$\uparrow$ \\
\toprule
&DCP~\cite{he2010single} & 16.62 & 0.818 & 14.01 & 0.760 
\\
&DehazeNet~\cite{cai2016dehazenet} & 19.82& 0.821 & 19.12 & 0.840 \\
&MSCNN~\cite{ren2016single} & 19.84 & 0.833 & 14.01 & 0.510 \\
&AOD-Net~\cite{li2017aod} & 32.33 & 0.950 & 27.17 & 0.898 \\
&GFN~\cite{ren2016single} & 22.30 & 0.880 & - & - \\
&GCANet~\cite{chen2019gated} & 30.23 & 0.980 & - & - \\ 
\textbf{Dehazing}&GridDehazeNet~\cite{liu2019griddehazenet} & 32.16 & 0.984 & 23.29 & 0.930 \\
\textbf{Task}&MSBDN~\cite{dong2020multi} & 33.67 & 0.985 & 22.99 & 0.850 \\
&PFDN~\cite{dong2020physics} & 32.68 & 0.976 & - & - \\
&FFA-Net~\cite{qin2020ffa} & 36.39 & 0.989 & \underline{26.96} & \underline{0.950} \\
&AECR-Net~\cite{wu2021contrastive} & 37.17 & 0.990 & - & - \\
&DehazeF-B~\cite{song2022vision} & 37.84 & \textbf{0.994} & - & - \\
&MAXIM-2S~\cite{tu2022maxim} & \underline{38.11} & \underline{0.991} & - & -  \\
&\textbf{GridFormer} & \textbf{42.34}& \textbf{0.994} & \textbf{33.27} & \textbf{0.986} \\
\bottomrule
\end{tabular}}
\label{tab:dehazing}
\end{table}
\begin{table}[t]
\centering
\scriptsize
\caption{Desnowing results on Snow100K-S/L. Bold and underlined fonts denote best and second-best results, respectively.
}
\setlength{\tabcolsep}{3pt}
\renewcommand{\arraystretch}{1.12}
\scalebox{0.85}{\begin{tabular}{c||c|cc|cc}
\hline
\rowcolor{mygray}
&& \multicolumn{2}{c}{Snow100K-S~\cite{liu2018desnownet}} & \multicolumn{2}{c}{Snow100K-L~\cite{liu2018desnownet}}  \\ 
\rowcolor{mygray}
\multirow{-2}*{Type}&\multirow{-2}*{Method} & PSNR$\uparrow$ & SSIM$\uparrow$ & PSNR$\uparrow$ & SSIM$\uparrow$ \\
\toprule
&SPANet~\cite{wang2019spatial} & 29.92 & 0.8260 & 23.70 & 0.7930 
\\
&JSTASR~\cite{chen2020jstasr} & 31.40 & 0.9012 & 25.32 & 0.8076 \\
&RESCAN~\cite{li2018recurrent} & 31.51 & 0.9032 & 26.08 & 0.8108 \\
\textbf{Desnowing}&DesnowNet~\cite{liu2018desnownet} & 32.33 & 0.9500 & 27.17 & 0.8983 \\
\textbf{Task}&DDMSNet~\cite{zhang2021deep} & 34.34 & 0.9445 & 28.85 & 0.8772 \\
&SnowDiff$_{64}$~\cite{ozdenizci2023restoring} & \underline{36.59} & \underline{0.9626} & \underline{30.43} & \underline{0.9145} \\ 
&SnowDiff$_{128}$~\cite{ozdenizci2023restoring} & 36.09 & 0.9545 & 30.28 & 0.9000 \\
&\textbf{GridFormer} & \textbf{38.89}& \textbf{0.9698} & \textbf{33.09} & \textbf{0.9340} \\
\midrule
&All-in-One \cite{li2020all} & - & - & 28.33 & 0.8820 \\
&TransWeather \cite{valanarasu2022transweather} & 32.51 & 0.9341 & 29.31 & 0.8879 \\
\textbf{Multi-weather}&Restormer~\cite{zamir2021restormer} & 36.08 & 0.9591 & 30.28 & 0.9124 \\
\textbf{Restoration}&WeatherDiff$_{64}$~\cite{ozdenizci2023restoring} & 35.83 & 0.9566 &
30.09 & 0.9041 \\
&WeatherDiff$_{128}$~\cite{ozdenizci2023restoring} & 35.02 & 0.9516 & 29.58 & 0.8941 \\
&\textbf{GridFormer-S} & \underline{36.68} & \underline{0.9602}& \underline{30.78} & \underline{0.9167} \\
&\textbf{GridFormer} & \textbf{37.46} &\textbf{0.9640} & \textbf{31.71} & \textbf{0.9231} \\
\bottomrule
\end{tabular}}
\label{tab:desnowing}
\end{table}
\subsection{Experimental Results}
\textbf{Dehazing results}. 
We perform image dehazing on different datasets to evaluate the performance of GridFormer. We compare the performance of GridFormer with various methods, including traditional prior-based methods, CNN-based methods, and recent transformer-based methods. Table~\ref{tab:dehazing} shows the quantitative results in terms of PSNR and SSIM. It shows that GridFormer achieves the best performance on the indoor subset of SOTS regarding all metrics. In particular,  GridFormer obtains a significant gain of $4.23$ dB in PSNR compared to the second-best method MAXIM-2S~\cite{tu2022maxim}.

We further compare the performance on the more challenging Haze4K dataset, which includes more realistic images from both indoor and outdoor scenarios. GridFormer obtains the best performance in terms of all metrics on this dataset as well. Fig.~\ref{fig:dehazing-visuals} provides a visual comparison for the SOTS indoor dataset. The recovered images by GridFormer contain finer details and are closer to the ground truth.

\begin{figure*}[t]
\begin{center}
	\includegraphics[width=\textwidth]{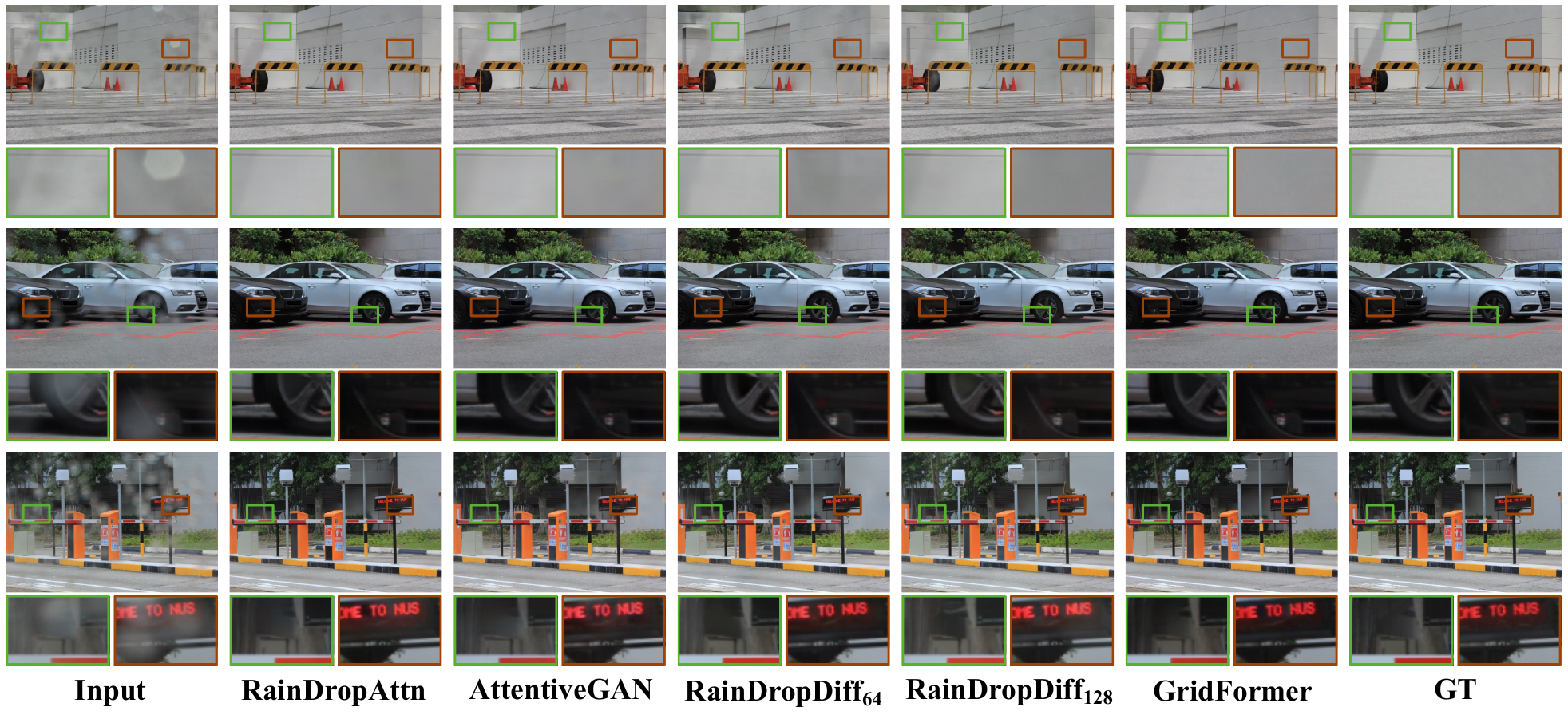}
	\caption{Raindrop removal results on RainDrop test set. From left to right are the input images, results of RaindropAttn~\cite{quan2019deep}, AttentiveGAN~\cite{qian2018attentive}, RainDropDiff$_{64}$~\cite{ozdenizci2023restoring}, RainDropDiff$_{128}$~\cite{ozdenizci2023restoring}, our GridFormer, and ground truth images, respectively. \textbf{Zoom in for details}.}
	\label{fig:raindrop-visuals}
 \end{center}
\end{figure*}

\begin{figure*}[t]
\begin{center}
	\includegraphics[width=\textwidth]{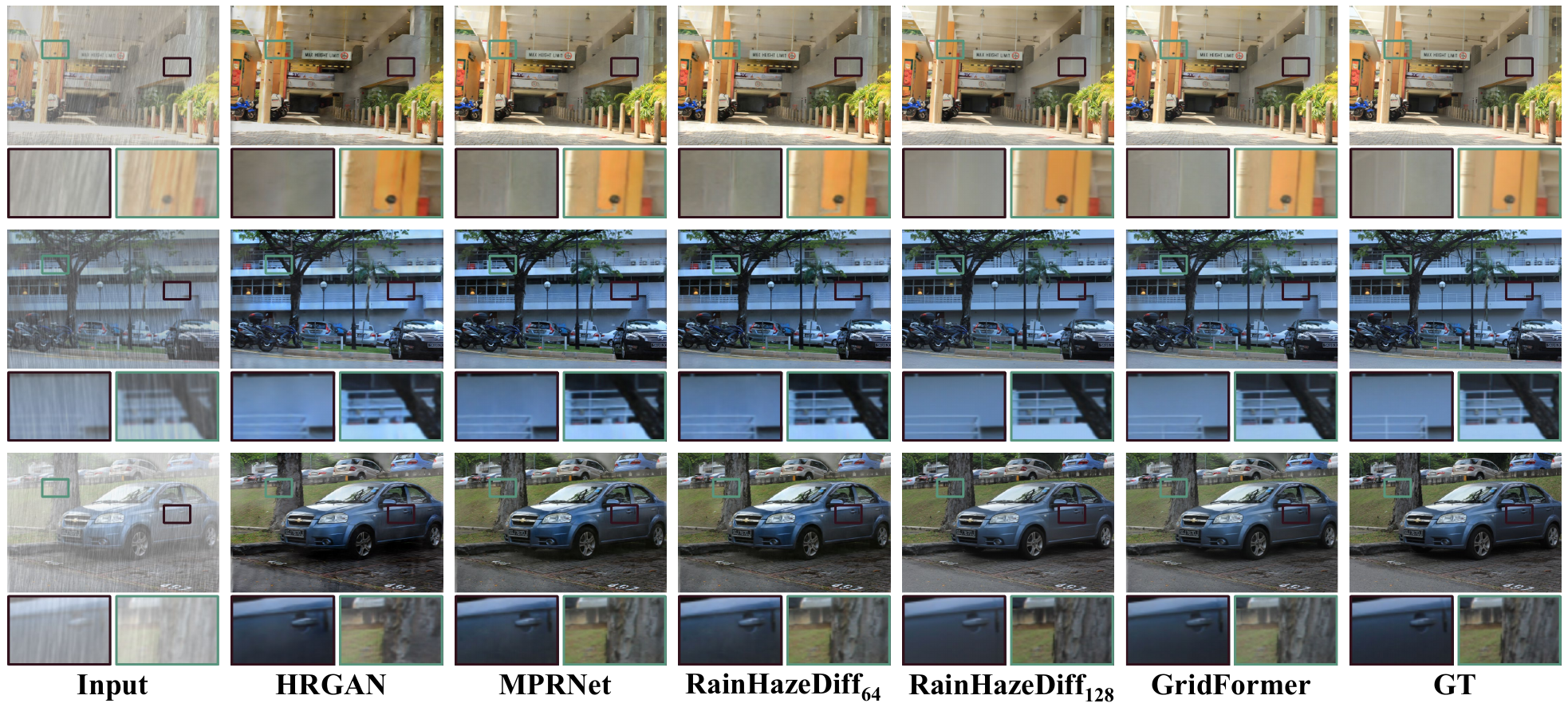}
	\caption{Visual results of deraining \& dehazing  on Outdoor-Rain test set. From left to right are the input images, results of HRGAN~\cite{li2019heavy}, MPRNet~\cite{zamir2021multi}, RainHazeDiff$_{64}$~\cite{ozdenizci2023restoring}, RainHazeDiff$_{128}$~\cite{ozdenizci2023restoring}, our GridFormer, and ground truth images, respectively. \textbf{Zoom in for details}.}
	\label{fig:deraining-and-dehazing-visuals}
 \end{center}
\end{figure*}

\textbf{Desnowing results}.
We evaluate the desnowing performance on the public Snow100K dataset~\cite{liu2018desnownet}. The test set is divided into three subsets according to the particle size: Snow100K-S, Snow100K-M, and Snow100K-L. We select Snow100K-S and Snow100K-L for testing. Table~\ref{tab:desnowing} shows the quantitative results. On the Snow100K-S subset, GridFormer outperforms the diffusion-based method SnowDiff$_{64}$~\cite{ozdenizci2023restoring} by $2.3$ dB and by $0.0072$ in terms of PSNR and SSIM. As for the most difficult Snow100K-L subset,  GridFormer still gains an improvement of $2.66$ dB and $0.0195$ in terms of PSNR and SSIM compared to the second-best method SnowDiff$_{64}$. Fig.~\ref{fig:desnowing-visuals-s} and Fig.~\ref{fig:desnowing-visuals-l} provide the visual comparisons, showing that GridFormer 
is effective in removing image corruption due to snow while producing perceptually pleasing results.

\begin{figure*}[t]
\begin{center}
	\includegraphics[width=\textwidth]{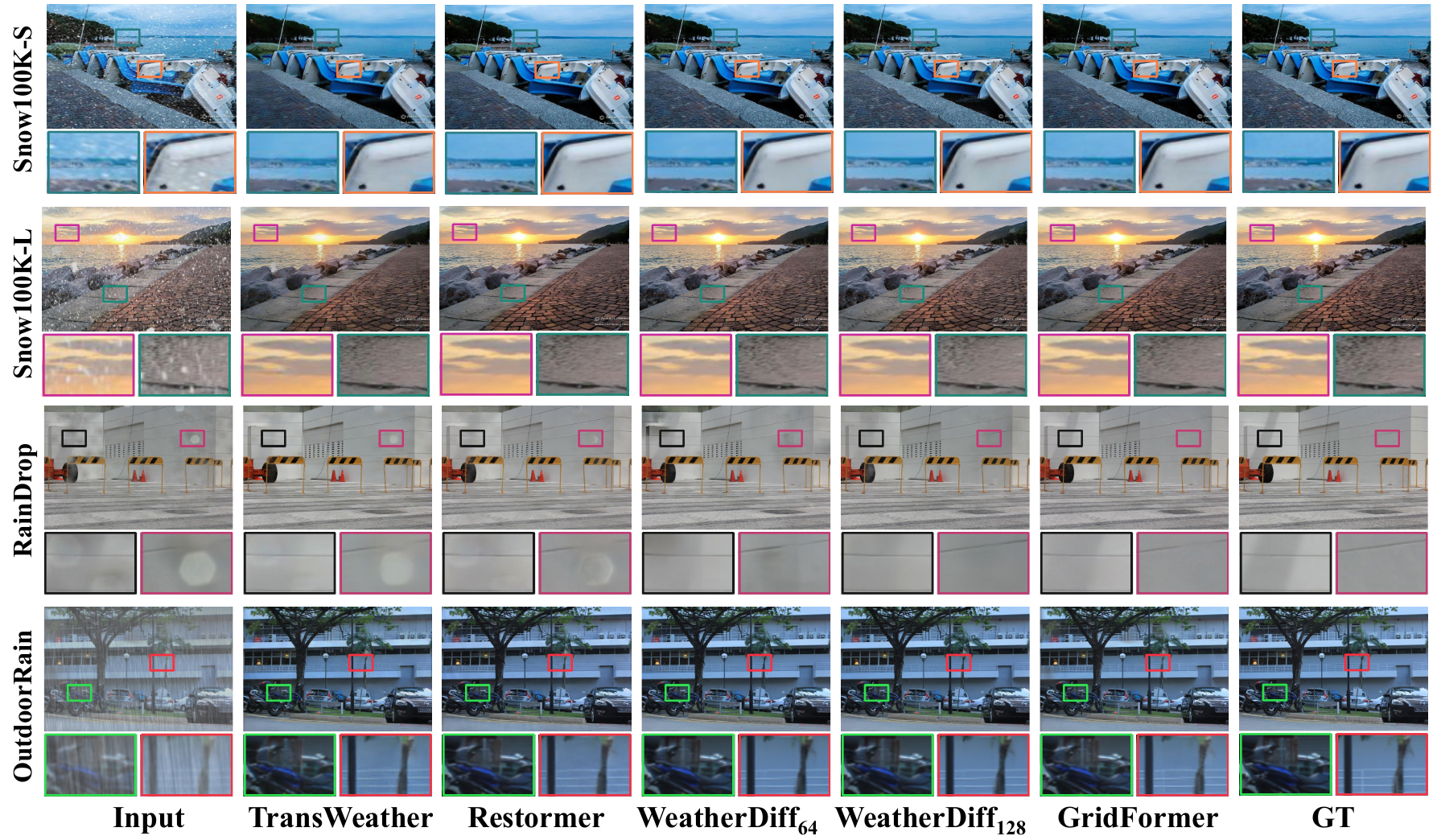}
	\caption{Multi-weather Restoration comparison on Snow100K-S, Snow100K-L, RainDrop, Outdoor-Rain datasets. From left to right are the input images, results of TransWeather~\cite{valanarasu2022transweather}, Restormer~\cite{zamir2021restormer}, WeatherDiff$_{64}$~\cite{ozdenizci2023restoring}, WeatherDiff$_{128}$~\cite{ozdenizci2023restoring}, our GridFormer, and ground truth images, respectively. \textbf{Zoom in for details}.}
	\label{fig:allweather-visuals}
 \end{center}
\end{figure*}

\begin{table}[t]
\centering
\scriptsize
\caption{RainDrop removal results on RainDrop test set. Bold and underlined fonts denote best and second-best results, respectively.}
\setlength{\tabcolsep}{8pt}
\renewcommand{\arraystretch}{1.2}
\scalebox{0.80}{\begin{tabular}{c||c|cc}
\hline
\rowcolor{mygray}
&& \multicolumn{2}{c}{RainDrop~\cite{qian2018attentive}}  \\ 
\rowcolor{mygray}
\multirow{-2}*{Type}&\multirow{-2}*{Method} & PSNR$\uparrow$ & SSIM$\uparrow$ \\
\toprule
&pix2pix~\cite{isola2017image} & 28.02 & 0.8547 \\
&DuRN~\cite{liu2019dual} & 31.24 & 0.9259 \\
&RaindropAttn~\cite{quan2019deep} & 31.44 & 0.9263 \\
\textbf{RainDrop}&AttentiveGAN~\cite{qian2018attentive} & 31.59 & 0.9170 \\
\textbf{Removal}&IDT~\cite{xiao2022image} & 31.87 & 0.9313 \\
&RainDropDiff$_{64}$~\cite{ozdenizci2023restoring} & 32.29 & \textbf{0.9422} \\ 
&RainDropDiff$_{128}$~\cite{ozdenizci2023restoring} & \underline{32.43} & 0.9334 \\
&\textbf{GridFormer} & \textbf{32.92} & \underline{0.9400} \\
\midrule
&All-in-One~\cite{li2020all} & \underline{31.12} & 0.9268 \\
&TransWeather~\cite{valanarasu2022transweather} & 30.17 & 0.9157 \\
\textbf{Multi-weather}&Restormer~\cite{zamir2021restormer} & 30.91 & 0.9282 \\
\textbf{Restoration}&WeatherDiff$_{64}$~\cite{ozdenizci2023restoring} & 29.64 & \underline{0.9312} \\
&WeatherDiff$_{128}$~\cite{ozdenizci2023restoring} & 29.66 & 0.9225 \\
&\textbf{GridFormer-S} & 31.02 & 0.9301   \\
&\textbf{GridFormer} & \textbf{32.39} &\textbf{0.9362}  \\
\bottomrule
\end{tabular}}
\label{tab:draindrop_removal}
\end{table}

\textbf{RainDrop removal results}.
In Table~\ref{tab:draindrop_removal} we present the quantitative results for raindrop removal on the RainDrop dataset. For an extensive comparison, we compare GridFormer with seven different methods: pix2pix~\cite{isola2017image}, DuRN~\cite{liu2019dual}, RaindropAttn~\cite{quan2019deep}, AttentiveGAN~\cite{qian2018attentive}, IDT~\cite{xiao2022image}, RainDropDiff$_{64}$~\cite{ozdenizci2023restoring}, and RainDropDiff$_{128}$~\cite{ozdenizci2023restoring}. The results show that  GridFormer is competitive. Specifically, GridFormer achieves the best performance in terms of PSNR and achieves almost the same level of performance as the state-of-the-art method RainDropDiff$_{64}$~\cite{ozdenizci2023restoring} in terms of SSIM with a difference of $0.0022$.  
A visual comparison of the results on RainDrop is provided in Fig.~\ref{fig:raindrop-visuals}. It shows that our method can remove raindrops successfully and generate realistic images.

\begin{table}[t]
\centering
\scriptsize
\caption{Image deraining \& dehazing results on Outdoor-Rain test set. The MACs of each model is measured on $256\times256$ image.}
\setlength{\tabcolsep}{6pt}
\renewcommand{\arraystretch}{1.1}
\scalebox{0.80}{\begin{tabular}{c||c|cc|c}
\hline
\rowcolor{mygray}
&& \multicolumn{2}{c|}{Outdoor-Rain~\cite{li2019heavy}} & Overhead \\ 
\rowcolor{mygray}
\multirow{-2}*{Type}&\multirow{-2}*{Method} & PSNR$\uparrow$ & SSIM$\uparrow$ & Param/MACs\\
\toprule
&CycleGAN~\cite{zhu2017unpaired} & 17.62 & 0.6560 &7.84M/42.38G \\
&pix2pix~\cite{isola2017image} & 19.09 & 0.7100 &54.41M/18.15G\\
&HRGAN~\cite{li2019heavy} & 21.56 & 0.8550& 25.11M/34.93G \\
\textbf{Deraining \&}&PCNet~\cite{jiang2021rain} & 26.19 & 0.9015 &627.56K/268.45G \\
\textbf{Dehazing}&MPRNet~\cite{zamir2021multi} & 28.03 & 0.9192 &3.64M/148.55G \\
&RainHazeDiff$_{64}$~\cite{ozdenizci2023restoring} & \underline{28.38} & \textbf{0.9320} &82.92M/475.16G\\ 
&RainHazeDiff$_{128}$~\cite{ozdenizci2023restoring} & 26.84 & 0.9152&85.56M/263.45G\\
&\textbf{GridFormer} &\textbf{28.49} & \underline{0.9213} &30.12M/251.35G\\
\midrule
&All-in-One~\cite{li2020all} & 24.71 & 0.8980 &44.00M/12.26G\\
&TransWeather~\cite{valanarasu2022transweather} & 28.83 & 0.9000 &21.90M/5.64G\\
&Restormer~\cite{zamir2021restormer} & 30.21 & 0.9208&26.10M/140.99G \\
\textbf{Multi-weather}&WeatherDiff$_{64}$~\cite{ozdenizci2023restoring} & 29.64 & 0.9312 &82.92M/475.16G\\
\textbf{Restoration}&WeatherDiff$_{128}$~\cite{ozdenizci2023restoring} & 29.72 & 0.9216&85.56M/263.45G \\
&\textbf{GridFormer-S} & \underline{30.48} & \underline{0.9313}&14.83M/133.24G
 \\
&\textbf{GridFormer} & \textbf{31.87} &\textbf{0.9335} &30.12M/251.35G \\
\bottomrule
\end{tabular}}
\label{tab:deraining_dehazing}
\end{table}

\textbf{Deraining and dehazing results}.
For the image deraining and dehazing task, we conduct experiments on the Outdoor-Rain dataset~\cite{li2019heavy}. This dataset has $9,000$ pairs of images for training, and $750$ pairs for testing, where degraded images are synthesized considering the rain and haze scenes simultaneously. The comparisons between  GridFormer and other state-of-the-art methods are reported in Table~\ref{tab:deraining_dehazing}.  GridFormer outperforms other competitors in terms of PSNR, and ranks second place regarding SSIM. More specifically, GridFormer achieves $0.11$ dB and $0.46$ dB improvement in terms of PSNR when compared to RainHazeDiff$_{64}$~\cite{ozdenizci2023restoring} and MPRNet~\cite{zamir2021multi}. Fig.~\ref{fig:deraining-and-dehazing-visuals} shows the visual comparison, indicating that GridFormer can handle haze and rainfall scenarios well at the same time, and generate vivid results. 

\begin{table}[t]
    \centering
    \scriptsize
    \caption{Cross-dataset evaluation. Models are trained only on the All-weather dataset and directly applied to the Rain100L and Test100 benchmark datasets. Bold and underlined fonts denote the best and second-best results, respectively.}
    \scalebox{1}{
    \begin{tabular}{c|cc|cc}
        \hline
        \rowcolor{mygray}
        & \multicolumn{2}{c}{Rain100L~\cite{yang2017deep}} & \multicolumn{2}{c}{Test100~\cite{zhang2019image}}  \\ 
        \rowcolor{mygray}
        \multirow{-2}*{Method} & PSNR$\uparrow$ & SSIM$\uparrow$  & PSNR$\uparrow$ & SSIM$\uparrow$  \\
        \toprule
        TransWeather   & 30.33 & 0.9365 & 24.20 & 0.8317 \\
        Restormer &27.08 & 0.8432 & 23.28  &  0.7136\\
        WeatherDiff64 & 27.46 & 0.8534 & 23.13 & 0.7091  \\
        WeatherDiff128 & 27.56 & 0.8552 & 23.26 & 0.7255 \\
        \textbf{GridFormer-S} & \underline{33.21} & \underline{0.9541} & \underline{27.10} & \underline{0.8713}   \\
        \textbf{GridFormer} & \textbf{34.24}& \textbf{0.9649} & \textbf{29.26} & \textbf{0.8912} \\
        \bottomrule
    \end{tabular}}
\label{tab:deraining_cross}
\end{table}

\begin{figure}
\resizebox{0.48\textwidth}{!}{
    \begin{tabular}{ccc}
        \includegraphics[width=0.32\linewidth]{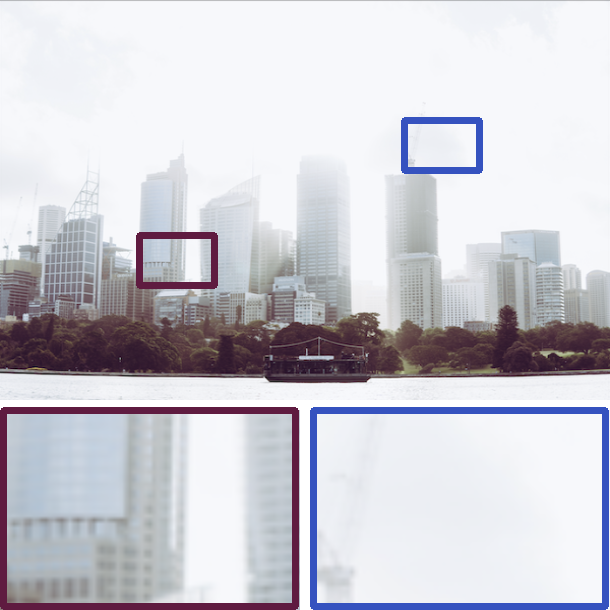} &
        \includegraphics[width=0.32\linewidth]{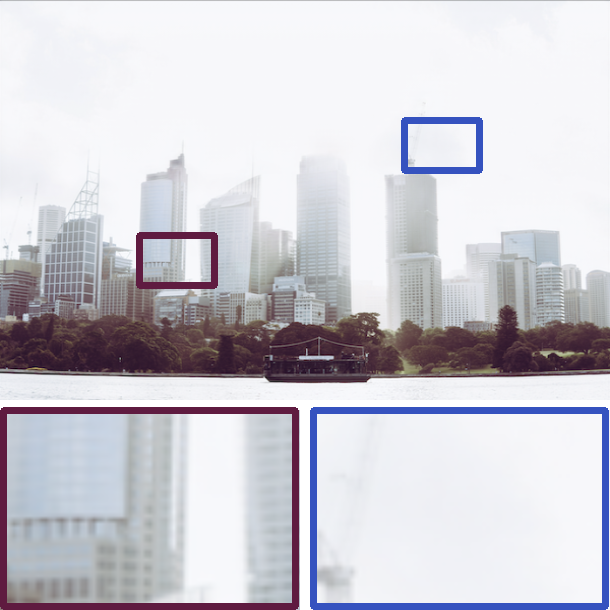} &
        \includegraphics[width=0.32\linewidth]{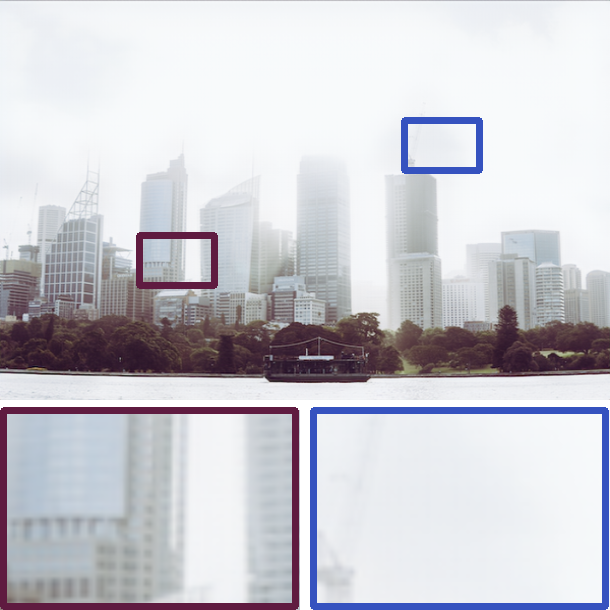}   \\
        \textbf{Input} &
        \textbf{Restormer} &
        \textbf{TransWeather}  \\
        \includegraphics[width=0.32\linewidth]{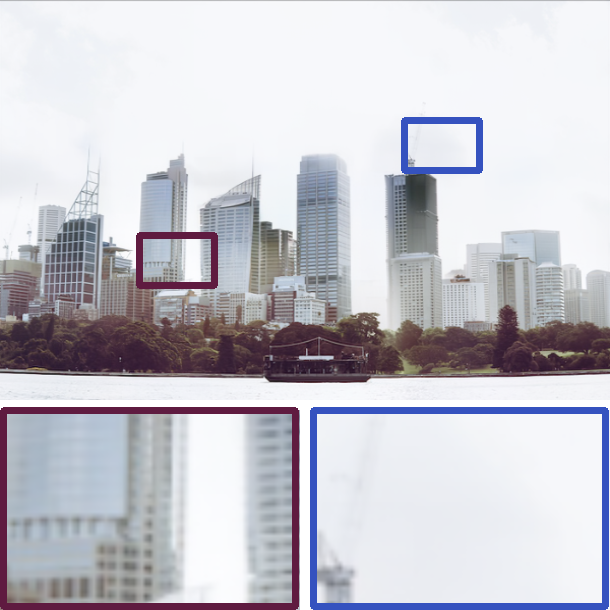} &
        \includegraphics[width=0.32\linewidth]{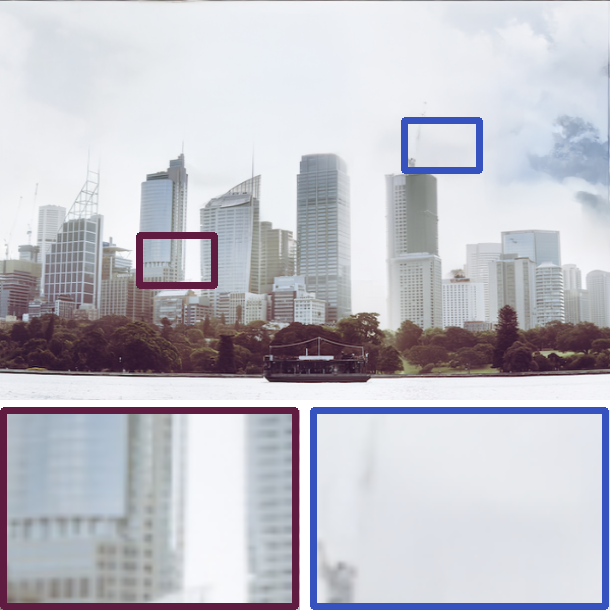} &
        \includegraphics[width=0.32\linewidth]{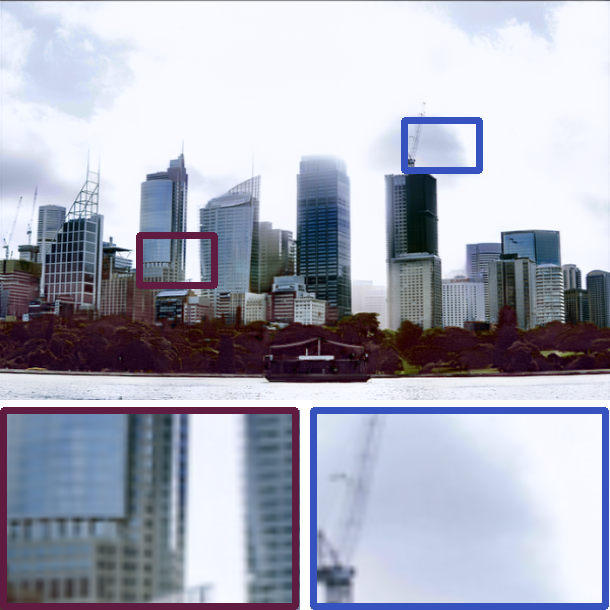}   \\
        \textbf{WeatherDiff$_{64}$}&
        \textbf{WeatherDiff$_{128}$} &
        \textbf{GridFormer}  \\
    \end{tabular}}
\caption{Exemplar results on the real-world image. }
    \label{fig:rw-comparison}
\end{figure}

\begin{figure*}[t]
	\centering
	 \includegraphics[width=0.95\textwidth]{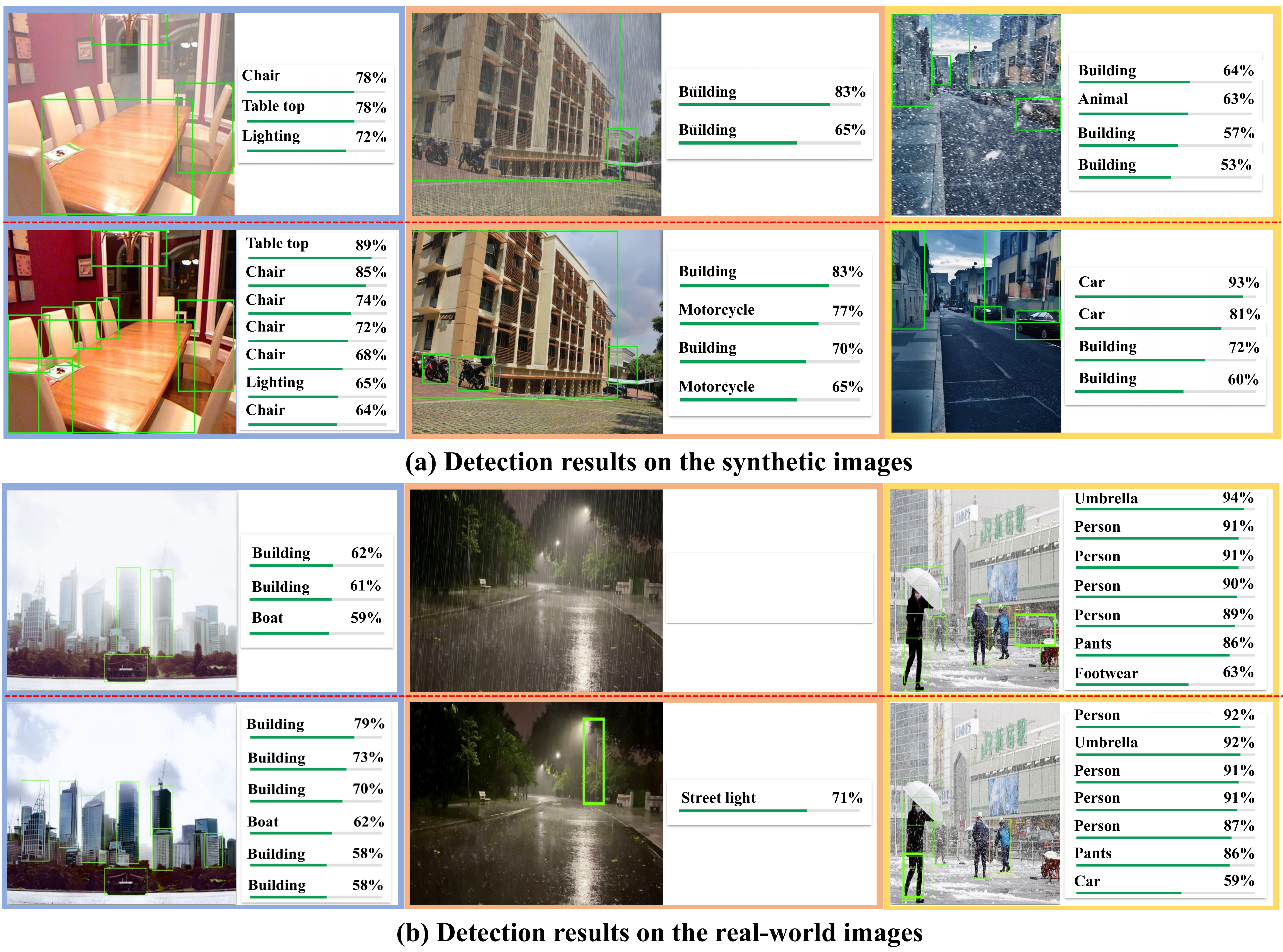}
 \caption{In each sub-image, the top images are captured in haze, rain, and snow weather conditions. The bottom images are recovered by our GridFormer. We report the detection confidences of these images, which shows that our GridFormer as a pre-processing tool benefits the task of object detection.}
	\label{fig:detection}
\end{figure*}

\textbf{Multi-weather restoration results}.
We further explore the potential of GridFormer for multi-weather restoration. Specifically, we first train our model on the mixed dataset sampled from Snow100K~\cite{liu2018desnownet}, Raindrop~\cite{qian2018attentive}, and Outdoor-Rain~\cite{li2019heavy} datasets. Then, we evaluate our model on the Snow100k-S/L test sets, the RainDrop test dataset, and the Outdoor-Rain test dataset. We choose four representative multi-weather restoration methods for comparison: All-in-One network~\cite{li2020all} is a CNN-based method, TransWeather is based on transformers, WeatherDiff$_{64}$ and WeatherDiff$_{128}$ are diffusion models. Table~\ref{tab:desnowing}, \ref{tab:draindrop_removal}, and \ref{tab:deraining_dehazing} summarize the quantitative results. GridFormer achieves the best performance in all weather conditions. We present visual comparisons in Fig.~\ref{fig:allweather-visuals}. Images produced by GridFormer exhibit fewer artifacts and are closer to ground truth compared to other methods. In an additional experiment, we set $C=32$ in the grid head to construct a tiny variant of the network called GridFormer-S for comparison. The results show that our methods achieve competitive results with less complexity and parameters, see Table~\ref{tab:deraining_dehazing}. 

\textbf{Cross-dataset evaluation}. To further verify the models' performance across different datasets, we conduct a cross-dataset evaluation on different SOTA methods. To be specific, the models (\ie TransWeather, Restormer, WeatherDiff$_{64}$, WeatherDiff$_{128}$, GridFormer-S, and GridFormer) are trained on the All-weather dataset, and then directly applied to the specific deraining datasets Rain100L~\cite{yang2017deep} and Test100~\cite{zhang2019image} for testing. Experimental results in Table~\ref{tab:deraining_cross} show that our GridFormer-S and GridFormer outperform other approaches.

\begin{table}[t]
    \centering
    \caption{Memory consumption and inference time of different methods evaluated on $512 \times 512$ resolution images.} 
    \setlength{\tabcolsep}{9pt}
\scalebox{0.9}{
    \begin{tabular}{c|ccc}
        \toprule[1pt]
        Methods & Platform& Memory (MB) & time (ms) \\       
        \hline 
        Restormer & PyTorch & 30031.50 & 321.6       \\
        WeatherDiff$_{64}$  & PyTorch & 6307.53  & 101232.1    \\
        WeatherDiff$_{128}$& PyTorch & 7941.03 & 133557.7     \\
        \hline \textbf{GridFormer-S}  &PyTorch & 20793.94 &  165.0 \\
        \textbf{GridFormer} &PyTorch & 28461.94 & 259.1           \\
        \bottomrule[1pt]
    \end{tabular}
    }
    \label{tab:efficient}
\end{table}
\textbf{Performance in real-world scenarios}. To further verify the effectiveness of the proposed method in real-world scenarios, we conduct a qualitative comparison experiment on the real-world hazy image from the Internet. The comparison result is shown in Fig.~\ref{fig:rw-comparison}. Compared with current state-of-the-art methods, our method effectively removes the haze and produces a clear result. The result shows that our method outperforms the current methods in real-world scenarios.

\textbf{Efficiency comparison}. We also analyze the efficiency of our models. Table~\ref{tab:efficient} displays the comparison results of different methods in terms of the memory consumption and inference time for $512 \times 512$ resolution. Specifically, we choose top three SOTA methods (\ie Restormer, WeatherDiff$_{64}$, and WeatherDiff$_{128}$) for comparison. Compared with other SOTA methods, our GridFormer-S exhibits the highest inference time. In addition, GridFormer-S and GridFormer are competitive in terms of memory consumption.

\begin{figure*}[t]
	\centering
	 \includegraphics[width=0.95\textwidth]{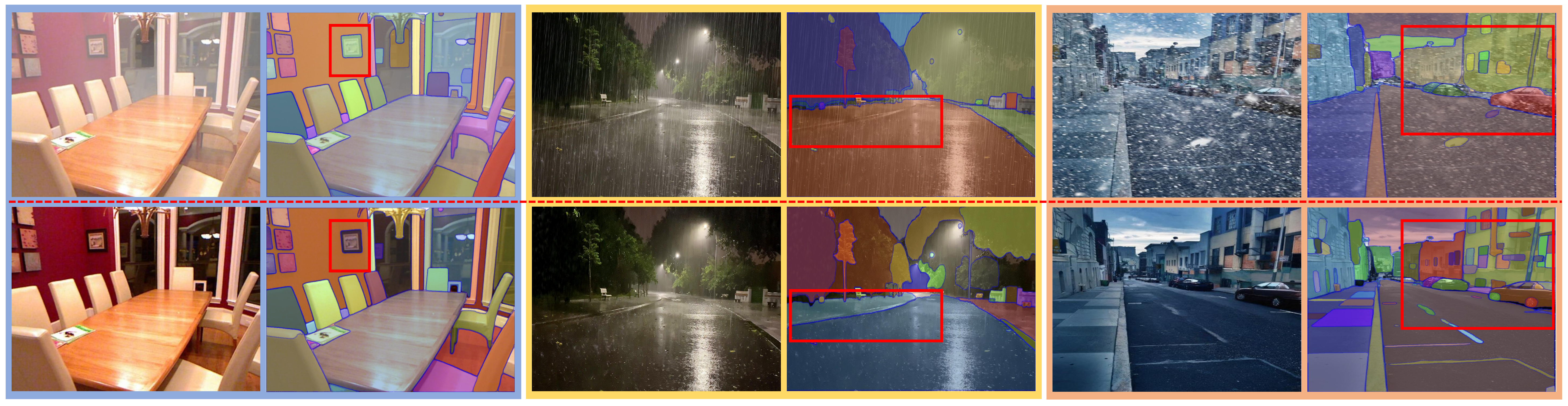}
 \caption{The top images are captured in haze, rain, and snow weather conditions. The bottom images are recovered by our GridFormer. We show the segmentation results of these images, which demonstrates that our GridFormer as a pre-processing tool benefits the task of image segmentation.}
	\label{fig:segmentation}
\end{figure*}
\begin{figure*}[t]
	\centering
	 \includegraphics[width=0.95\textwidth]{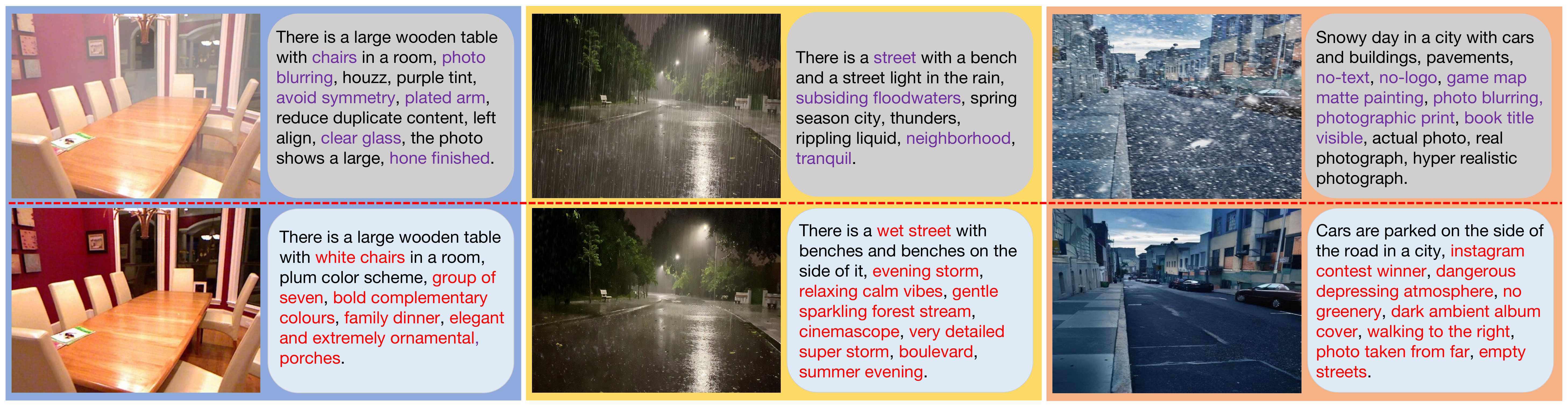}
 \caption{In each sub-image, the top images are captured in haze, rain, and snow weather conditions. The bottom images are recovered by our GridFormer. We show the image caption results of these images, which shows that our GridFormer as a pre-processing tool benefits the task of image caption.}
	\label{fig:caption}
\end{figure*}

\subsection{Application}
Image restoration in adverse weather conditions can enhance the image content, which can be easily incorporated into other high-level vision tasks. As a result, we investigate the potential of GridFormer in improving the performance of object detection, image segmentation, and image caption algorithms when dealing with adverse weather scenes. 
In the case of object detection, we consider both synthetic and real-world images. Fig.~\ref{fig:detection} shows detection results, where we use Google Vision API for object detection. We observe that haze, rain, and snow greatly reduce the detection accuracy, that is, increased missed detection, higher false detection, and lower detection confidence. In contrast, the detection
accuracy and confidence of the images recovered by GridFormer show significant improvement over those of weather-degraded images.
Fig.~\ref{fig:segmentation} showcases the segmentation results, utilizing the state-of-the-art Segment Anything Model~\cite{kirillov2023segment} for image segmentation. Our restoration results demonstrate an enhancement in segmentation accuracies, indicating that GridFormer effectively facilitates subsequent segmentation performance. Lastly, Fig.~\ref{fig:caption} presents the image caption results using the BLIP model~\cite{li2022blip}. These results show that the BLIP model can generate detailed captions by utilizing our restoration results, further validating the effectiveness of our GridFormer.

\begin{table}[!t]
\centering
\scriptsize
\caption{Ablation studies on the proposed compact-enhanced self-attention (CESA). FS, CS, and LE refer to feature sampling, channel split, and local enhancement operations in CESA respectively. MACs are measured on $256\times256$ images.}
\setlength{\tabcolsep}{5.5pt}
\begin{tabular}[t]{@{}c@{}}
\begin{tabular}{ccc|c|c|c}
\arrayrulecolor{black}\toprule

 \multirow{2}*{\scriptsize{FS}}  &  \multirow{2}*{\scriptsize{CS}} &  \multirow{2}*{\scriptsize{LE}} &  \multirow{2}*{\scriptsize{Param/MACs}} &  RainDrop &  SOTS-Indoor  \\ \cline{5-6} 
    &   &   &   & \scriptsize{PSNR/SSIM} &  \scriptsize{PSNR/SSIM}  \\
\midrule
\xmark  & \xmark  & \xmark  & 38.08M/322.26G & 29.93/0.8710 & 36.89/0.9631 \\ \arrayrulecolor{lightgray}\hline
 \cmark &  \xmark & \xmark  &34.91M/237.79G& 30.82/0.9012 & 37.81/0.9831  \\ \arrayrulecolor{lightgray}\hline
 \cmark & \cmark &  \xmark &  26.88M/227.65G& 31.98/0.9294 & 40.51/0.9921
   \\ \arrayrulecolor{lightgray}\hline
 \cmark & \cmark & \cmark & 30.12M/251.35G& 32.57/0.9365 & 41.84/0.9932
  \\ 
\arrayrulecolor{black}\bottomrule
\end{tabular}

\end{tabular}
\label{tab:ablation_CESA}
\end{table}

\subsection{Ablation Study}\label{ablation_study}
We conduct extensive ablation studies to verify the proposed compact-enhanced self-attention, residual dense transformer block, grid structure, and used loss functions. Specifically, we conduct ablation studies on the tasks of raindrop removal and dehazing to analyze the performance of GridFormer. For each model, we train it for $2\times10^{5}$ iterations using a batch size of $12$ on the RainDrop dataset~\cite{qian2018attentive} and the ITS dataset~\cite{li2018benchmarking} respectively. Subsequently, we assess the performance of each model on the testing sets of the RainDrop dataset and the testing SOTS-Indoor dataset. The detailed results are presented as follows.

\textbf{A. Compact-enhanced self-attention.} We verify the impact of feature sampling, channel split, and local enhancement operations in compact-enhanced self-attention. Table~\ref{tab:ablation_CESA} shows the comparison results. After applying feature sampling (FS) and channel split (CS) operations respectively, the model achieves $0.89$ dB and $1.16$ dB improvements in the RainDrop dataset ($0.92$ dB and $2.70$ dB improvements in the SOTS-Indoor dataset), while the computational complexity is significantly reduced. Using local enhancement (LE) operations, the performance gains on the RainDrop and SOTS-Indoor datasets are $0.59$ dB and $1.33$ dB respectively. The ablation study results suggest the effectiveness of these operations. We also conduct an additional ablation study on the RainDrop dataset to verify the effectiveness of exchanging the Values for feature interaction and fusion in our compact-enhanced self-attention. 
Specifically, we focus on exchanging only the Queries between $\mathbf{z}_{1}$ and $\mathbf{z}_{2}$ to investigate its impact on performance. The results of this experiment demonstrate that exchanging the Queries results in a PSNR value of $31.81$, which is inferior to the outcome achieved by exchanging the Values ($32.57$). These ablation results show the effectiveness of the Value exchange in enhancing interaction and feature fusion, thereby contributing to improved restoration performance. Furthermore, in our model, the choice of the step parameter $r$, as described in formula 3, indeed impacts the model's computational complexity and performance. Thus, we evaluate the effect of different settings of $r$ in the GridFormer. Table~\ref{tab:ablation_r} shows that our model with the $[4, 2, 2]$ setting achieves a better trade-off between computation cost and performance.

\begin{table}[!t]
\centering
\scriptsize
\caption{Ablation studies on the different settings of $r$ in three rows of GridFormer layers, where $r$ indicates the stride of the average pooling operation in feature sampling of the proposed compact-enhanced transformer layer. $[4,2,2]$ denotes $r$ is set as 4, 2, and 2 in three rows of GridFormer layers, respectively.}
\setlength{\tabcolsep}{3.5pt}
\scalebox{0.95}{\begin{tabular}[t]{@{}c@{}}
\begin{tabular}{c|c|c|c}
\arrayrulecolor{black}\toprule

\multirow{2}*{Different Settings of $r$}  &  \multirow{2}*{\scriptsize{Param/MACs}} &  RainDrop &  SOTS-Indoor  \\ \cline{3-4} 
   &    & \scriptsize{PSNR/SSIM} &  \scriptsize{PSNR/SSIM}  \\
\midrule
$[2,2,2]$ & 29.53M/213.01G & 30.93/0.8991 & 33.81/0.9631 \\ \arrayrulecolor{lightgray}\hline
$[4,4,4]$ &48.63M/356.57G& 31.62/0.9112 & 40.31/0.9901  \\ \arrayrulecolor{lightgray}\hline
$[4,2,2]$ &30.12M/251.35G& 31.98/0.9294 & 40.51/0.9921\\
\arrayrulecolor{black}\bottomrule
\end{tabular}

\end{tabular}}
\label{tab:ablation_r}
\end{table}
\begin{table}[!t]
\centering
\scriptsize
\caption{Ablation studies on the proposed residual dense transformer block (RDTB). DC, LF, and LSC denote dense connection, local fusion with $1\times1$ convolution, and local skip connection in RDTB respectively. MACs are measured on $256\times256$ images.}
\setlength{\tabcolsep}{5pt}
\begin{tabular}{ccc|c|c|c}
\arrayrulecolor{black}\bottomrule
 \multirow{2}*{\scriptsize{DC}}  &  \multirow{2}*{\scriptsize{LF}} &  \multirow{2}*{\scriptsize{LSC}} & \multirow{2}*{\scriptsize{Param/MACs}} &   RainDrop &  SOTS-Indoor   \\ \cline{5-6}

&   &   &   & \scriptsize{PSNR/SSIM} &  \scriptsize{PSNR/SSIM}  \\ 
\midrule
 \xmark & \xmark  &  \xmark & 27.99M/253.57G & 31.48/0.9187 & 40.32/0.9901
 \\ \arrayrulecolor{lightgray}\hline
 \cmark & \xmark  &  \xmark &32.78M/284.87G& 32.07/0.9284 & 40.45/0.9912   \\ \arrayrulecolor{lightgray}\hline
 \cmark & \cmark & \xmark  &  30.12M/251.35G& 32.05/0.9298 & 40.87/0.9926  \\ \arrayrulecolor{lightgray}\hline
 \cmark & \cmark & \cmark & 30.12M/251.35G& 32.57/0.9365 & 41.84/0.9932
  \\ 
\arrayrulecolor{black}\bottomrule
\end{tabular}
\label{tab:ablation_rstb}
\end{table}
\begin{table}[t]
	\footnotesize
	\center
 \caption{Ablation study on different gird configurations. $r$ and $c$ denote the numbers of rows and columns of the model.}
  \scalebox{0.75}{\begin{tabular}{c|c|c|c|c|c}
		\hline
		\multicolumn{2}{c|}{Grid Setting} &\multicolumn{2}{c|}{Overhead}  &RainDrop & SOTS-Indoor\\
		\cline{1-6}
		$r$& $c$  & Param (M) &  MACs (G)&  PSNR/SSIM  &  PSNR/SSIM     \\
		\hline
		\multirow{4}*{$r=1$} & $c=3$ & 0.81 & 51.24	& 30.07/0.9163 & 38.02/0.9879	\\
		\cline{2-6}
		&	$c=4$ & 1.01 & 64.01 & 30.44/0.9198  & 38.21/0.9880 \\
           \cline{2-6}
            &	$c=5$ &  1.21& 76.77 & 30.54/0.9199  & 38.61/0.9885\\
            \cline{2-6}
		&	$c=6$ & 1.41 & 89.54&30.62/0.9205	& 38.78/0.9891 \\
		\cline{1-6}
		\multirow{4}*{$r=2$}	&	$c=3$	&2.64 & 80.48&31.21/0.9280 & 39.68/0.9900	\\
		\cline{2-6}
  		&	$c=4$	&3.50 &103.98 & 31.23/0.9281 & 39.73/0.9901\\
		\cline{2-6}
		&	$c=5$	&4.51 & 129.52 &31.37/0.9294 & 40.21/0.9915\\
		\cline{2-6}
		&	$c=6$	&5.37 & 153.01 & 31.58/0.9311	& 40.50/0.9918	\\
		\cline{1-6}
		\multirow{4}*{$r=3$} &	$c=3$	& 13.96& 125.38&31.67/0.9339 & 40.79/0.9918	\\
		\cline{2-6}
          & $c=4$	& 19.09& 166.01 &31.73/0.9356 & 41.13/0.9929	\\
		\cline{2-6}
		&	$c=5$	& 24.99 &210.72 &31.89/0.9359 & 41.01/0.9926	\\
		\cline{2-6}
		&	$c=6$	& 30.12& 251.35&\textbf{32.57/0.9365} & \textbf{41.84/0.9932} \\
		\cline{1-6}
		$r=4$	&	$c=6$	& 150.86& 410.09&32.05/0.9361 & 40.85/0.9921	\\
		\hline
	\end{tabular}}
		
		\label{tab:row-column}
\end{table}

\textbf{B. Residual dense transformer block.} To demonstrate the effectiveness of the proposed residual dense transformer block, we conduct ablation studies by considering the following three factors: (1) dense connections (DC), (2) local fusion with $1\times1$ convolution (LF), and (3) local skip connection (LSC). Specifically, we analyze the different models by progressively adding these components. The results are shown in Table~\ref{tab:ablation_rstb}. We observe that each component improves the performance, where dense connections contribute the most.

\textbf{C. Exploring different configurations in the grid structure of GridFormer.} 
To comprehensively understand the impact of GridFormer's grid structure, we have conducted ablation experiments involving variations in the number of rows and columns. Each row within our GridFormer framework corresponds to a distinct scale, while the columns in the grid fusion module act as conduits that facilitate the exchange of information across diverse scales. This grid structure profoundly influences the information interchange that occurs among the grid units within the Grid Fusion module. In our study, we have systematically altered the number of rows, ranging from $1$ to $4$, while maintaining columns at values of $3$, $4$, $5$, and $6$. The results with different configurations are shown in Table~\ref{tab:row-column}. By increasing $r$ and $c$, the performance is improved, and the overhead gradually becomes complex. The model performance achieves its maximum for $r=3$ and $c=6$. Thus, we select these values in our final model. 

\textbf{D. Other GridFormer components.}  
The skip connection from input images and the perceptual loss also contribute to improving the performance. Without the skip connection from the input image, the PSNR value would decrease from $32.57$ dB to $31.85$ dB on the testing set of the RainDrop dataset. Training GridFormer without the perceptual loss results in a PSNR of $32.72$ dB on the testing set of the RainDrop dataset.

\section{Limitations and Future Work}\label{sec:limitations}
As a new backbone, GridFormer has achieved better performance than previous methods in image restoration under adverse weather conditions, but it still has space for improvement. For example, using the pre-trained strategy~\cite{chen2021pre} or the contrastive learning technique~\cite{wu2021contrastive} on our GridFormer can further explore its performance potential. In addition, we fuse multi-scale features with simple weighted attention~\cite{zheng2022t,wang2022ultra}. We can improve this fusion by designing special modules using sophisticated attention mechanisms~\cite{qin2020ffa,song2022vision}. Finally, GridFormer is evaluated in the image scenery, and we are still exploring whether it can handle the video restoration problem. In the future, it is also an important direction to extend our GridFormer to deal with video restoration in adverse weather conditions.

\section{Conclusion} \label{sec:conclusion}
In this paper, we propose GridFormer, a unified Transformer architecture for image restoration in adverse weather conditions. It adopts a grid structure to facilitate information communication across different streams and makes full use of the hierarchical features from the input images. In addition, to build the basic layer of GridFormer, we propose a compact-enhanced transformer layer and integrate it in a residual dense manner, which encourages feature reuse and enhances feature representation. Comprehensive experiments show that GridFormer significantly surpasses state-of-the-art methods, producing good results on both weather-specific and multi-weather restoration tasks.

\section*{Acknowledgement}
This work was supported in part by the National
Natural Science Foundation of China (GrantNo. 62372223, 62372480), in part by the Guangdong Basic and Applied Basic Research Foundation (No. 2023A1515012839), in part by Shenzhen Science and Technology Program (No. JSGG20220831093004008), in part by China Mobile Zijin Innovation Insititute (No. NR2310J7M).

\section*{Data Availability Statement}
The datasets generated during and/or analyzed during the current study are available in the 
WeatherDiffusion repository, with the link as https://github.com/IGITUGraz/WeatherDiffusion.
{\small
\bibliographystyle{spmpsci}
\bibliography{egbib}
}

\end{document}